\documentclass[lettersize,journal]{IEEEtran}
\usepackage{amsmath,amsfonts}
\usepackage{algorithmic}
\usepackage{algorithm}
\usepackage{array}
\usepackage{textcomp}
\usepackage{stfloats}
\usepackage{url}
\usepackage{verbatim}
\usepackage{subcaption}
\usepackage{graphicx}
\usepackage{cite}
\usepackage{hyperref}
\usepackage[numbers]{natbib}
\usepackage{color}
\usepackage{cleveref}
\usepackage{multirow} 
\usepackage{booktabs} 
\newcommand{\qyp}[1]{\textcolor{black}{#1}}
\begin{document}

\title{A Survey on Explainable Reinforcement Learning: Concepts, Algorithms, and Challenges}

\author{Yunpeng~Qing,
        Shunyu~Liu,
        Jie~Song,
        Yang~Zhou,
        Kaixuan~Chen,
        Huiqiong~Wang$^\dagger$,
        Mingli~Song
\thanks{$^\dagger$Corresponding author.}
\thanks{Y.~Qing, J.~Song, and H.~Wang are with the College of Computer Science and Technology, Zhejiang University, Hangzhou 310027, China (e-mail: qingyunpeng@zju.edu.cn, sjie@zju.edu.cn, huiqiong\_wang@zju.edu.cn).}
\thanks{S.~Liu, is with the College of Computing and Data Science, Nanyang Technological University
, 639798, Singapore  (e-mail: liushunyu@ntu.edu.cn).}
\thanks{Y.~Zhou is with School of Software Technology, Zhejiang University, Hangzhou 310027, China (e-mail: imzhouyang@zju.edu.cn).}
\thanks{K.~Chen and M.~Song are with State Key Laboratory of Blockchain and Security, Zhejiang University, Hangzhou 310027, China  (e-mail: chenkx@zju.edu.cn, brooksong@zju.edu.cn).}
}
        % <-this % stops a space
% \thanks{Manuscript received Feburary 15, 2025.}}

\markboth{Review}%
{Qing \MakeLowercase{\textit{et al.}}: A Survey on Explainable Reinforcement Learning: Concepts, Algorithms, and Challenges}


\maketitle
\begin{abstract}
        Reinforcement Learning~(RL) is a popular machine learning paradigm where intelligent agents interact with the environment to fulfill a long-term goal. Driven by the resurgence of deep learning, Deep RL~(DRL) has witnessed great success over a wide spectrum of complex control tasks.
        Despite the encouraging results achieved, the deep neural network-based backbone is widely deemed as a black box that impedes practitioners to trust and employ trained agents in realistic scenarios where high security and reliability are essential. 
        To alleviate this issue, a large volume of literature devoted to shedding light on the inner workings of the intelligent agents has been proposed, by constructing intrinsic interpretability or post-hoc explainability.  
        In this study, we conducted a comprehensive review of existing work on explainable RL (XRL) and introduced a new classification scheme, categorizing previous work into several main categories, namely, agent model explanation, reward explanation, state explanation, and task explanation, and further dividing them below.
        % We also review and highlight RL methods that conversely leverage human knowledge to promote learning efficiency and performance of agents while this kind of method is often ignored in XRL field. 
        Some challenges and opportunities in XRL are discussed. This survey intends to provide a high-level summarization of XRL and to motivate future research on more effective XRL solutions. Corresponding open source codes are collected and categorized at \url{https://github.com/Plankson/awesome-explainable-reinforcement-learning}.
\end{abstract}
\begin{IEEEkeywords}Reinforcement Learning, Explainability.\end{IEEEkeywords}   
\section{Introduction}\label{sec1}
% Reinforcement learning~\cite{RL} is inspired by human trial-and-error paradigm~\cite{noble1957human} in which interacting with the environment is a common way for human learning without the guidance of others~\cite{kjellstrom2010tracking}. 
% Through interactions, humans acquire experiential knowledge regarding cause and effect, action outcomes, and goal attainment within the environment. This acquired experience is subsequently leveraged implicitly to formulate our mental models ~\cite{yampolskiy2012artificial,williamson1999mental,powers2006advisor}, enabling us to navigate and resolve the encountered tasks efficiently~\cite{borgman1986user,rowe1995measuring}. Similarly, RL autonomously learns from interacting with environments to purposefully understand environment dynamics and influence future events. Technically, RL learns to map from environment state to action so as to maximize the cumulative reward~\cite{stauffer2016components}.

Reinforcement Learning (RL)~\cite{RL} is a computational framework for training autonomous agents to purposefully understand environmental dynamics and influence future events through sequential interactions with the environment. This paradigm is inspired by human trial-and-error learning mechanisms, where interaction with the environment serves as a fundamental approach to learning without external guidance\cite{kjellstrom2010tracking,yampolskiy2012artificial,williamson1999mental,powers2006advisor}. In Technical, RL learns to map environmental states to actions to maximize cumulative rewards~\cite{stauffer2016components}.

In recent years, the fast development of deep learning~\cite{bengio2017deep,sze2017efficient} promotes the fusion of deep learning and reinforcement learning. Therefore, Deep Reinforcement Learning~(DRL)~\cite{DQN,PPO,A3C,SAC,TD3} has emerged as a new RL paradigm. 
% With the powerful representation capability of the deep neural network~\cite{ansuini2019intrinsic,yosinski2014transferable,goldfeld2018estimating}, DRL has achieved considerable performance in many domains~\cite{chen2019attention,liu2019deep,duan2019deep,AlphaGo,Dota,lin2018efficient,challita2019interference,liu2023CIA}. Particularly in game-based tasks such as AlphaZero~\cite{AlphaGo} and OpenAI Five~\cite{Dota}, DRL methods have achieved remarkable success by outperforming human professional players.
With the powerful representation capability of the deep neural network~\cite{ansuini2019intrinsic,yosinski2014transferable,goldfeld2018estimating}, DRL has achieved considerable performance in many domains, such as games~\cite{AlphaGo, Dota} and robotic tasks~\cite{chen2019attention,liu2023CIA, qinga2po, qing2025bitrajdiff, zhou2023centralized}.
However, in complex real-world scenarios like autonomous driving~\cite{fayjie2018driverless,wang2018deep,chen2019model,hoel2019combining,wang2017formulation} and power system dispatch~\cite{{chen2024powerformer,xu2024temporal,lin2020deep,yang2021dynamic,liu2023MAM,liu2023PAC}}, both high performance and user-oriented explainability should be taken into account to ensure security and reliability. Therefore, the lack of explainability in DRL is one of the main bottleneck for employing DRL in the real world.

The opacity of conventional deep reinforcement learning (DRL) stems from the inherent complexity of deep neural networks (DNNs)~\cite{he2016deep}, where high-dimensional parameter spaces and nonlinear transformations impede traceability. This architectural intricacy prevents the identification of features driving decisions or the mechanisms processing them~\cite{zahavy2016graying}, rendering DRL models functionally opaque ``black boxes"~\cite{jaunet2020drlviz}.
Such opacity engenders two critical challenges. The first one is the trust barrier~\cite{ivchyk2024overcoming}, where agent decisions conflict with human intuition without explainable justifications. For example, In autonomous navigation~\cite{prasetya2020navigation}, abrupt route deviations~(e.g., avoiding unperceived congestion) may confuse users despite rational objectives. And the second one is the knowledge integration limits~\cite{han2020improving,rosenfeld2018leveraging}. While encoding human expertise, the inability to map DRL representations to human-explainable concepts hinders effective integration~\cite{zhang2019leveraging,zhang2019faster,guan2020explanation,guan2021widening,silva2021encoding}.
Explainable AI (XAI) techniques, such as saliency maps~\cite{object-saliency-map,perturbation-based-saliency} and symbolic reasoning~\cite{STATE,Ration-learning, Query-based}, partially address these issues by linking model internals to human-understandable rationales. Although various XAI techniques have been adopted to computer vision~\cite{x_f_r_1,x_f_r_2,x_f_r_3,x_f_r_4} and natural language processing field~\cite{x_nlp_1,x_nlp_3,x_nlp_4}, their direct adaptation to DRL remains challenging due to temporal decision dependencies and partial observability of RL tasks~\cite{Survey_2}.

In the field of eXplainable Reinforcement Learning~(XRL), many preliminary studies and surveys~\cite{XRL_3,heuillet2021explainability,Survey_4,Survey_2,broad-xai,self-interpretable-2,milani2023explainable} have made effort to construct the XRL model and gained certain achievements in producing explanations. 
Early surveys~\cite{heuillet2021explainability,self-interpretable-2} directly adopted XAI's intrinsic vs. post-hoc dichotomy, categorizing XRL by explanation timing. While effectively distinguishing model-transparent approaches like decision trees~\cite{LMUT} from post-hoc analyzers such as casual model~\cite{AIM}, this framework fails to address RL-specific components like environment dynamics and reward model~\cite{Survey_2}. 
Subsequent works decomposed explanations along RL structural elements, \cite{Survey_2} partitioned methods by target MDP components into state features, environment dynamics, and agent mechanism; \cite{XRL_3} simply focused on agent preferences and goal influences; and \cite{milani2023explainable} proposed policy, feature, and MDP-centric categories. 
However, these taxonomies suffer from overlapping boundaries. For example, programmatic policies~\cite{verma2019imitation} simultaneously address reward decomposition~(feature-level) and long-term goals (policy-level), defying singular categorization. 
\cite{broad-xai} introduced a causal architecture that spans from perceptions to dispositions. Although theoretically comprehensive, current implementations~\cite{prasetya2020navigation} operationalize only perception-action causation, representing a much simpler form of the causal framework.
In conclusion, current XRL research faces three systemic challenges:~
(1)~Absence of Standardization:~Despite diverse conceptual frameworks mentioned above, the field lacks consensus on foundational elements, including the rigorous definitions of explainability in RL and integral evaluation protocols for XRL methods.
% (1)~Although many of them proposed their own understanding for XRL~\cite{explainability_1,XRL_1,explainability_2,XRL_3}, the XRL field currently is still lacking standard criteria, especially for its definition and evaluation approaches.
(2)~Taxonomy-Application Misalignment:~Existing categorization schemes exhibit two key mismatches. The intrinsic/post-hoc dichotomies directly taken from the XAI taxonomy inadequately capture RL components, while the current component-based XRL taxonomy creates overlapping categories. This necessitates a unified taxonomy based on MDP formalism, explicitly distinguishing explanations of different MDP component targets from various XRL technical approaches.
% (2)~Current taxonomies do not align with the existing XRL methods. One part of the work employs the XAI taxonomy to categorize existing research, ignoring the significant components of RL like reward and policy, which is unable to cover all the XRL methods comprehensively. Meanwhile, traditional XAI focuses on machine learning predicting with labeled or unlabeled data, which is quite different from RL about interaction with the environment to maximize rewards. This difference in the target leads XAI taxonomy to be inappropriate for XRL.
% The other part introduces novel taxonomies for XRL but falls short in accurately summarizing and distinguishing existing XRL methods. The XRL community requires a more concise taxonomy directly derived from the RL paradigm to categorize various XRL works clearly. Such a taxonomy is crucial for comprehending and advancing XRL techniques.
% (3)~Human-Centric Knowledge Integration Gap:~In the XRL literature, the role of human participation has often been overlooked. However, several papers have recently attempted to integrate human prior knowledge, such as human-annotated trajectories~\cite{garg2022lisa} and online collaborations involving human command~\cite{gao2023towards}, into the XRL learning process. The results of these studies strongly indicate that incorporating human prior knowledge is an effective approach for achieving both high explainability and performance.
(3)~Human-Centric Knowledge Integration Gap:~While recent work demonstrates the efficacy of human-RL cooperation like trajectory annotations~\cite{garg2022lisa} and real-time corrections~\cite{gao2023towards}, current XRL taxonomies fail to formalize knowledge integration pathways. Standardized frameworks are needed to highlight the effectiveness of human prior knowledge for both high explainability and performance.

To advance the further development of XRL, this survey makes a more comprehensive and specialized review on XRL concepts and algorithms. We first clarify the concepts of RL explainability, then we give a systematic overview of the existing evaluation metrics for XRL, encompassing both subjective and objective assessments.
We proposed a new taxonomy that categorizes current XRL works according to the central target of explanation: agent model, reward, state, and task, precisely capturing the central component in RL paradigm. 
Since making the whole RL paradigm explainable is currently difficult, all of the works turn to get partial explainability directly on components of RL paradigms. 
This taxonomy is much more specialized than the general coarse-grained intrinsic/post-hoc or global/local taxonomies in XAI, providing clearer distinctions among existing XRL methods and comprehensive illustration of RL decision-making process. Meanwhile, by assigning each method to a specific category aligning with its primary objective and specific implementation details, the taxonomy avoids ambiguity or confusion in the category process.
Meanwhile, given that there is currently only a small amount of research on human knowledge-integrated XRL and its importance, we make an attempt on summarizing these works and organizing them into our taxonomy. As we know, few researchers look into this field of integrating human knowledge into XRL.
Our work can be summarized below:
\begin{itemize}
  \vspace{-0.1cm}
        \item We give a clear definition of RL explainability by summarizing existing literature on explainable RL. What's more, we also propose a systematic evaluation architecture of XRL from objective and subjective aspects.
        \item To make up for the shortcomings of lacking RL-based architecture in XRL community, we propose a new RL-based taxonomy for current XRL works. The taxonomy is based on the explainability of different central target of the RL framework: agent model, reward, state, and task. The taxonomy can be viewed in Figure~\ref{Fig1}.
        \item 
        % Noticing that currently human knowledge-intergrated XRL is an emerging direction, based on our new XRL taxonomy we give a systematic review of these approaches that combines XRL frameworks with human prior knowledge to get higher performance and better explanation. 
        Recognizing that human knowledge-integrated XRL is an emerging research direction, we provide a systematic review of these approaches based on our new XRL taxonomy. This review illustrates how XRL frameworks incorporate human prior knowledge to enhance performance and improve explainability.
  \vspace{-0.05cm}
\end{itemize}

\begin{figure*}
  \centering
  \vspace{-0.3cm}
  \includegraphics[width=1.02\textwidth]{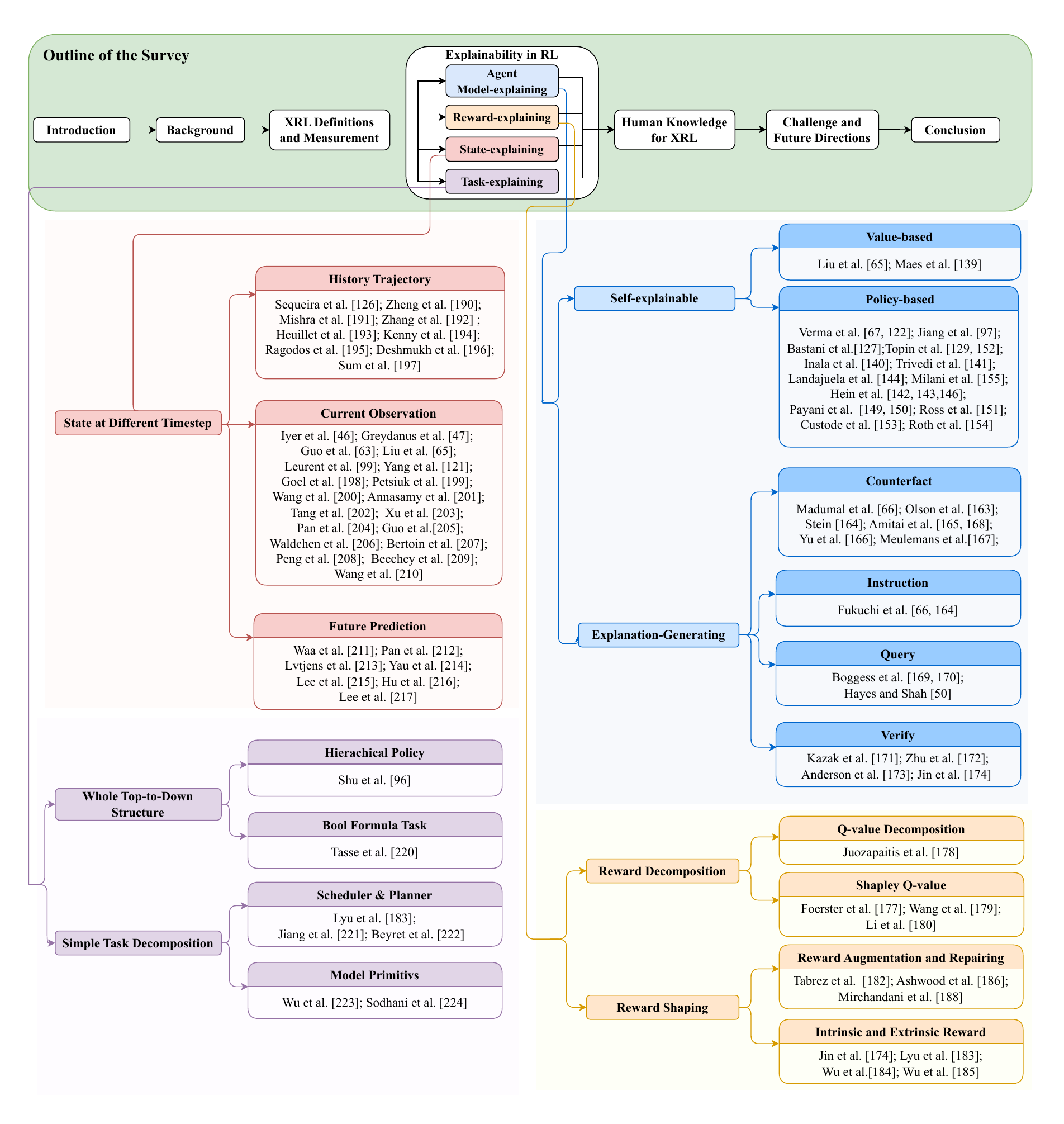}
  \caption{An overview of the survey.  We categorize existing explainable reinforcement learning~(XRL) approaches into four branches based on the explainability of different parts in RL process: agent model, reward, state, and task. The more fine-grained categorization will be discussed detailedly in later sections. Each category is demonstrated with a part of representative works in the figure with different colors.}
  \vspace{-0.3cm}
  \label{Taxonomy}
\end{figure*}

The remaining of this survey is organized as follows. In Section~\ref{sec:background}, we recall the necessary basic knowledge of reinforcement learning. Next, we discuss 
the definition of RL explainability as well as giving some possible evaluation aspects for explanation and XRL approaches in Section~\ref{sec::Explainable_RL_definitions_and_measurement}. 
In Section~\ref{sec::Explainability_in_RL}, we describe our categorization as well as provide works of 
each type and sub-type in detail, the abstract figure of our taxonomy can be viewed in Figure~\ref{Fig1}. Then we discuss XRL works that are combined with human knowledge according to our taxonomy in Section~\ref{sec::Human_knowledge_for_RL_paradigm}. 
After that, we summarize current challenges and promising future directions of XRL in Section~\ref{sec::Challenges and future directions for XRL}.
Finally, we give a conclusion of our work in Section~\ref{sec::Conclusion_and_future_work}. The structure of this paper and our taxonomy work is shown in Figure~\ref{Taxonomy}. 
\section{Background}
\label{sec:background}
 Reinforcement Learning paradigm considers the problem of how an agent interacts with the environment to maximize the cumulative reward, where the reward is a feedback signal according to the response action of the agent in different states. Concretely, the interaction process can be formalized as a Markov Decision Process~(MDP)~\cite{MDP}. An MDP is described as a tuple $M = \langle \mathcal{S},\mathcal{A},P,R,\gamma \rangle$, where $\mathcal{S}$ is the state space, $\mathcal{A}$ is the action space, $P: \mathcal{S} \times \mathcal{A} \times \mathcal{S} \to [0,1]$ is the state transition function, $R: \mathcal{S} \times \mathcal{A} \to \mathbb{R}$ is the reward function, and $\gamma \in [0,1]$ is a discount factor. At each discrete time step $t$, the agent observes the current state $s_t \in \mathcal{S}$ and chooses an action $a_t \in \mathcal{A}$. This causes a transition to the next state $s_{t+1}$ drawn from the transition function $P(s_{t+1}|s_t,a_a)$. Moreover, the agent can receive a reward signal $r_t$ according to the reward function $R(s_t,a_t)$. The core object of the agent is to learn an optimal policy $\pi^*$ that maximizes the expected discounted return $\mathbb{E}_{\pi}[G_t] = \mathbb{E}_{\pi}[\sum_{i=0}^{\infty}\gamma^i r_{t+i}]$. To tackle this problem, existing reinforcement learning methods can be categorized into two classes: value-based methods and policy-based ones.

\subsection{Value-based Methods}
The value-based methods~\cite{DQN} tend to assess the quality of a policy $\pi$ by the action-value function $Q^{\pi}$ defined as:
\begin{align}
    Q^\pi(s,a)=\mathbb{E}_{\pi}[\sum_{i=0}^{\infty}{\gamma^i r_{t+i}} | s_t=s,a_t=a],\nonumber
\end{align}
which denotes the expected discounted return after the agent executes an action $a$ at state $s$.
A policy $\pi^*$ is optimal if:
\begin{align}
Q^{\pi^*}(s,a) \geq Q^\pi(s,a), \forall{ \pi, s \in \mathcal{S}, a \in \mathcal{A}}.\nonumber
\end{align}
There is always at least one policy that is better than or equal to all other policies~\cite{RL}. All optimal policies share the same optimal action-value function defined as $Q^*$. It is easy to show that $Q^*$ satisfies the Bellman optimality equation:
\begin{align}
    Q^*(s,a) = \mathbb{E}_{s' \sim P(\cdot |s,a)}\left[R(s,a)+\gamma \max_{a'\in\mathcal{A}}Q^*(s',a')\right].\nonumber
\end{align}

To estimate the optimal action-value function $Q^*$, Deep $Q$-Networks~(DQN)~\cite{DQN} uses a neural network $Q(s,a; \theta)$ with parameters $\theta$ as an approximator. We optimize the model by minimizing the following temporal-difference~(TD) loss:
\begin{align}
    \mathcal{L}(\theta)=\mathbb{E}_{(s,a,r,s') \sim \mathcal{D}}\left[\left(y-Q(s,a;\theta)\right)^2\right],\nonumber
\end{align}
where $\mathcal{D}$ is the replay buffer of the transitions, $y = r+\gamma \max_{a'}Q(s',a';\theta^-)$ and $\theta^-$ represents the parameters of the target network. After the network converges, the final optimal policy can be obtained by a greedy policy $\pi(s) = \arg\max_{a\in \mathcal{A}} Q(s,a;\theta)$. Due to the encouraging results accomplished by DQN, several follow-up works~\cite{Double-DQN,Dueling-DQN,DQN-modify,DQN-POMDP,PER,Distributional-DQN,Rainbow-DQN,ES_DQN,DQN_update,DQN_update1} progressively enlarged the family of DQN and has recently demonstrated extraordinary capabilities in multiple domains~\cite{DQN_use2,DQN_use3,DQN_use4,DQN_use5}. 
However, while these value-based methods can handle high-dimensional observation spaces, they are restricted to problems with discrete and low-dimensional action spaces.

\subsection{Policy-based Methods}
To solve the problems with continuous and high-dimensional action spaces, policy-based methods have been proposed as a competent alternative. One of the conventional policy-based methods is stochastic policy gradient~(SPG)~\cite{RL}, which seeks to optimize a policy function $\pi_\phi:\mathcal{S} \times \mathcal{A} \to [0,1]$ with parameters $\phi$. SPG directly maximizes the expected discounted return as the objective $\mathcal{J}(\phi)=\mathbb{E}_{\pi_{\phi}}[\sum_{t=0}^{\infty}\gamma^t r_{t}]$. To update the policy parameters $\phi$, we can perform the gradient of this objective as follow:
\begin{align}
    \nabla_\phi \mathcal{J}(\pi_\phi) = \mathbb{E}_{s\sim \rho^{\pi} ,a\sim\pi_{\phi}}\left[\nabla_\phi \log \pi_{\phi}(a | s) Q^{\pi}(s,a)\right],\nonumber
\end{align}
where $\rho^{\pi}(s)$ is the state distribution and $Q^{\pi}(s,a)$ is the action value. To estimate the action value $Q^{\pi}(s,a)$, a simple and direct way is to use a sample discounted return $G$.
Furthermore, to reduce the high variance of the action-value estimation while keeping the bias unchanged, a general method is to subtract an estimated state-value baseline $V^\pi(s)$ from return~\cite{RL}. This yields the advantage function $A^{\pi}(s,a) = Q^{\pi}(s,a) - V^{\pi}(s)$, where an approximator $V(s;\theta)$ with parameters $\theta$ is used to estimate the state value. This method can be viewed as an actor-critic architecture where the policy function is the actor and the value function is the critic~\cite{TRPO,PPO,DPPO,GAE,A3C,AC_modify1,AC_modify2,AC_modify3}.

On the other hand, the policy in the actor-critic architecture can also be updated through the deterministic policy gradient~(DPG)~\cite{DPG,DDPG,TD3} for continuous control:
\begin{align}
    \nabla_\phi \mathcal{J}(\mu_\phi) = \mathbb{E}_{s\sim \rho^{\mu}}\left[\nabla_a Q^\mu(s,a)|_{a=\mu_{\phi}(s)} \nabla_\phi \mu_{\phi}(s)\right].\nonumber
\end{align}
where $\mu_{\phi}(s): \mathcal{S} \to \mathcal{A}$
% with parameters $\phi$ 
is a deterministic policy. Moreover, we directly instead approximate the action-value function $Q^\mu(s,a)$ with a parameterized critic $Q(s,a; \theta)$, where the parameters $\theta$ are updated using the TD loss analogously to the value-based case. By avoiding a problematic integral over the action space, DPG provides a more efficient policy gradient paradigm than the stochastic counterparts~\cite{DPG}.

\section{Explainable RL definitions and measurement}
\label{sec::Explainable_RL_definitions_and_measurement}
This section establishes the foundation for integrating explainability into frameworks. 
Although numerous studies in this field have attempted to establish a precise definition of explainable RL, neither standardized criteria nor a clear consensus have emerged within the research community. 
Moreover, many current studies treat explainability as a subjective perception that does not require rigorous analysis. This conceptual ambiguity impedes both the development of the XRL research and the standardized evaluation metrics for XRL frameworks. 
Following a comprehensive review of existing literature, we provide a detailed conceptual analysis of XRL and categorize current evaluation metrics for XRL systems.

\subsection{Definition of XRL}
In this section, we establish a unified definition of XRL, addressing inconsistencies in existing literature caused by divergent criteria proposed across scholarly works. 
A review of key studies reveals that many approaches~\cite{XRL_1, XRL_2} frame explainability through human-centric interactions. For instance, \cite{XRL_1} defines RL explainability as the degree to which humans can understand an agent decision, while \cite{XRL_2} asserts it as the degree to which humans can consistently predict a model output.
Concurrently, other works~\cite{self-interpretable-2, milani2023explainable} adopt the XAI taxonomy, distinguishing between interpretability~(self-explanatory models) and explainability~(user-oriented explanations), yet omit a comprehensive formal definition. 
Synthesizing these perspectives, XRL seeks to provide transparent explanations of decision-making processes in sequential contexts, enabling human understanding of both the process and outcomes, facilitating behavioral prediction, and ensuring reliable generation of user-aligned explanations. Based on the explanatory focus identified in surveyed studies, current XRL methods can be classified into the following categories:
\begin{itemize}
\item[(1)] \textit{Intrinsic Explainability via Agent Architecture}:
The internal agent architecture can be designed with inherent transparency. This form of explainability measures the extent to which the decision-making logic and inner mechanisms remain explainable throughout its operational lifecycle (training and deployment). Representative architectures include decision trees~\cite{Decision_tree}, hierarchical agents~\cite{STG}, and logic-based rule systems~\cite{NLRL_1}, among others.
\item[(2)] \textit{Extrinsic Explainability via Post-Hoc Analysis}:
This category involves generating supplementary explanations to post-decision to clarify the agent action outputs. These explanations identify factors influencing the agent action selection, such as state-input dependencies or feature contributions. Common techniques include saliency maps~\cite{huber2021local}, attention  distributions~\cite{leurent2019social}, \textit{etc}.
\end{itemize}
Intrinsic explainability is an inherent property established during the agent design phase. In contrast, extrinsic explainability requires not only a fully developed agent but also input data and its execution traces, rendering it a post-hoc characteristic. Thus, the XRL field is fundamentally structured around these two classes of explanations: intrinsic~(design-driven) and extrinsic~(execution-dependent).

\subsection{Evaluation Framework}
Having established a definition of explainability, we now turn to the evaluation of XRL. However, the field lacks standardized metrics for assessing the explainability of RL frameworks. While scholars have proposed preliminary frameworks, such as \cite{Interpretability_evaluation} introducing a three-tiered evaluation approach including application, human, and function-level criteria, and \cite{Metrics_for_XAI,Metrics_for_XAI_plus} developing quantifiable metrics for explainable AI, no unified methodology has gained broad acceptance. Building on these efforts, we summarize these contributions into a structured evaluation framework tailored to XRL paradigm:

\subsubsection{Subjective Assessment}
Subjective assessment evaluates XRL frameworks by analyzing human-generated mental models, which users construct to interpret the agent decision-making processes and structure~\cite{yampolskiy2012artificial,williamson1999mental,powers2006advisor}. By inversely assessing these mental models, researchers can gauge the effectiveness of explanations. Subjective evaluation seeks to quantify the accuracy of these mental representations; however, directly measuring internal cognitive states remains challenging. Current methodologies rely on indirect human feedback. Key metrics for subjective assessment include the following categories:
% Subjective assessment assesses explainable frameworks based on human feedback. Users receive the explanation and construct the mental model of how a person understands the model process and structure~\cite{yampolskiy2012artificial,williamson1999mental,powers2006advisor}. Therefore, conversely evaluating the mental model can be a feasible way to verify the effectiveness of the explanation. The subjective assessment aims to capture and measure the mental model of the user. However, it is hard to directly measure the mental model of human users in their minds. Current approaches to the mental model are all based on human feedback. Significant metrics for subjective assessment can be categorized as follows:
\begin{itemize}
\item [(1)] \emph{User Prediction~(S.UP)}:~This metric evaluates the alignment between human predictions and the RL agent actual decisions, serving as a proxy for the fidelity of the mental model derived from explanations. A quantitative way is to make the user predict the agent decision $a_\text{pred}$, compare to the real agent action $a_\text{RL}$ and calculate the hit rate $\sum_N|a_\text{pred}-a_\text{RL}|^2/N$, where N denotes the total number of trials~\cite{kay2016ish,ribeiro2016should,ribeiro2018anchors,nushi2018towards,bansal2019beyond}.
Questionnaires are also widely applied to quantify participants’ self-reported understanding of the agent decision logic and explanations~\cite{kim2018interpretability,kulesza2013too,lakkaraju2016interpretable,rader2015understanding}, often using Likert-scale responses to assess task-specific comprehension.

\item [(2)] \emph{User Confidence~(S.UC)}:~User confidence measures the perceived reliability of explanations, distinct from user prediction metrics that assume full reliance on RL explanations. It reflects the persuasiveness of explainability in fostering trust and actionable reliance on the agent decisions~\cite{bilgic2005explaining,gedikli2014should,lage2019human}. 
Structured questionnaires are widely employed to quantify confidence levels~\cite{lim2009assessing,coppers2018intellingo,lim2009and,berkovsky2017recommend,AIM}.
Many researchers~\cite{pu2006trust,nothdurft2014probabilistic} track the actions and intentions of users through questionnaires to measure their trust and reliance on the explanations.
% To assess trust and reliance, studies often analyze behavioral indicators (e.g., adherence to explanations) and self-reported intentions captured through surveys~\cite{pu2006trust,nothdurft2014probabilistic}.
Additionally, response time $\Delta t$ is leveraged as an implicit metric to quantify explanation complexity, where prolonged deliberation suggests reduced clarity~\cite{lim2009and,gedikli2014should}.

\item [(3)] \emph{Descriptiveness~(S.D)}:~Many XRL literature employs descriptive case studies, such as state feature visualizations~\cite{perturbation-based-saliency, RS-rainbow} and programmatic policy examples~\cite{verma2019imitation, PIRL}, to demonstrate explainability through illustrative scenarios. While such descriptions enhance persuasiveness by logically contextualizing agent decisions, they inherently lack quantitative validation. Notably, almost all XRL studies adopt this approach to assert explainability~\cite{LMUT,leurent2019social,liu2023curricular}, which positions descriptiveness as an informal, subjective metric due to its reliance on qualitative evidence rather than empirical measurement.
\end{itemize}

Subjective assessment leverages human feedback to implicitly gauge the efficacy of explanations. 
Advances in questionnaire design methodologies~\cite{hermans1970questionnaire,lietz2010research} now facilitate the development of structured questionnaires that systematically quantify S.UP and S.UC, making them a cornerstone of explainability evaluation~\cite{AIM,introspection}. 
However, these metrics remain vulnerable to inherent biases arising from subjective human interpretation, such as cultural predispositions or task familiarity. 
To ensure robustness, researchers must implement rigorous controls, such as diverse participant sampling and blinded evaluation protocols, to mitigate biases and ensure equitable assessment conditions.

\subsubsection{Objective Assessment}
Objective assessment focuses on quantitatively evaluating explanations using purely algorithmic outputs, thereby eliminating the need for human feedback.  These objective metrics provide quantitative evaluations of the agent effectiveness and avoid the potential biases introduced by subjective human judgment.
\begin{itemize}
\item [(1)] \emph{Decision Performance~(O.DP)}:~Decision performance quantifies the cumulative rewards $G_\pi$ achieved by an RL agent $\pi$. It is crucial not to sacrifice performance in favor of explainability. Consequently, O.DP is a mandatory metric for all XRL paradigms, ensuring explanatory mechanisms do not degrade agent effectiveness~\cite{PIRL,VIPER,leurent2019social,tang2021sensory,topin2021iterative, liu2023curricular}.

\item [(2)] \emph{Fidelity~(O.F)}:~Fidelity measures the measures faithfulness of explanations to the agent actual behavior, ascertaining the extent to which the explanation accurately portrays the agent behavior~\cite{Metrics_for_XAI_plus}. 
The quantification of fidelity varies depending on the employed methodologies. Concerning the intrinsic explanation via agent architecture, fidelity is assessed by measuring the disparity between the inexplainable policy $\pi_\text{RL}$ and the intrinsic explainable policy $\pi_\text{XRL}$, denoted as $D(\pi_\text{RL}, \pi_\text{XRL})$~\cite{liu2023fidelity}, where $D$ quantifies discrepancies in action distributions.
For extrinsic explanations via post-hoc analysis, fidelity can be assessed by perturbing states highlighted in explanations and measuring the change of returns~\cite{guo2021edge}.
\item [(3)] \emph{Robustness~(O.R)}: Robustness quantifies the stability of explanations under perturbations~\cite{kindermans2019reliability,sundararajan2017axiomatic}. 
The assessment of robustness involves evaluating and validating the generated explanation under different conditions.
Current evaluation methods add perturbation on state input~\cite{binder2016analyzing,nguyen2020model,binder2016analyzing} or model parameters~\cite{adebayo2018local,adebayo2018sanity,gevrey2003review}, utilizing the difference of generated explanation to determine whether the explanation remains stable.
\end{itemize}
Objective assessment provides a systematic framework for evaluating RL explainability by relying on algorithmic outputs. While it ensures reproducibility and mitigates human bias, its exclusion of human feedback may lead to imprecise measurement of explainability correctness. Meanwhile, current objective assessment metrics are limited to the specific type of explanation.
The field of XRL currently lacks universally accepted measurement methods, necessitating further research attention to address this gap.

\label{sec::Explainability_in_RL}
\begin{figure*}[!htb]
	\begin{subfigure}{0.245\textwidth}
		\centering
 		\includegraphics[width=\textwidth]{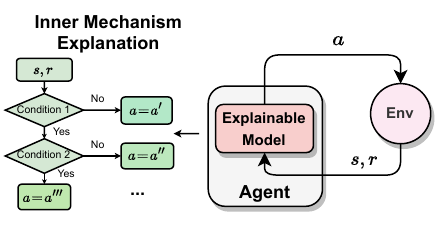}
		\caption{Agent Model-explaining}
    \label{Model}
	\end{subfigure}
	\begin{subfigure}{0.245\textwidth}
		\centering
		\includegraphics[width=\textwidth]{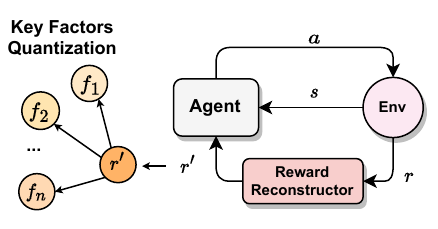}
		\caption{Reward-explaining}
    \label{Reward}
	\end{subfigure}
	\begin{subfigure}{0.245\textwidth}
		\centering
		\includegraphics[width=\textwidth]{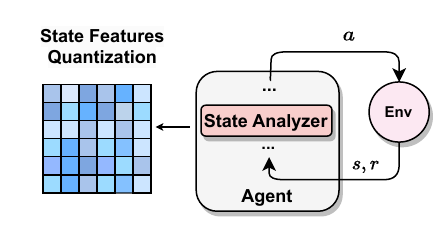}
		\caption{State-explaining}
    \label{State}
	\end{subfigure}
	\begin{subfigure}{0.245\textwidth}
		\centering
		\includegraphics[width=\textwidth]{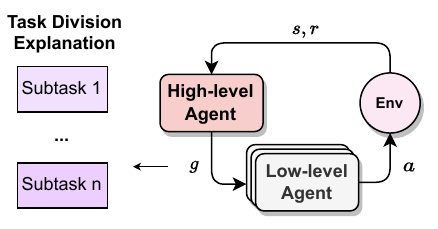}
		\caption{Task-explaining}
    \label{Task}
	\end{subfigure}
  \caption{Diagrams of different types of XRL frameworks. These diagrams illustrate how different types of XRL make different parts of the RL model produce explanations. (a) constructs the agent on an explainable model to illustrate the inner mechanism. (b) reconstructs reward function $r$ towards an explainable one $r'$, which is constructed by quantifying the quantitative impact of various key factors $\{w_i\}$. (c) adds a state analyzer submodule to quantify the influences of state features for each state input $s$.
  (d) gets an architectural level explainability in complex tasks by task division and subtask signal $g$. }

  \label{Fig1}
\end{figure*}
\section{Explainability in RL}

% We propose a taxonomic framework for eXplainable Reinforcement Learning~(XRL) grounded in the explainability targets of core RL components. Given the inherent complexity and computational intractability of full process explanation, current XRL approaches strategically prioritize specific interpretable dimensions while preserving algorithmic performance. Our framework organizes XRL methodologies according to four fundamental RL elements: state, reward, agent architecture, and task structure.

We construct the taxonomy of XRL based on the central target of explanation under the RL paradigm. 
We find that XRL researches focus on making certain aspects of the RL paradigm understandable while maintaining performance. The underlying model of RL can be segmented into several components, namely state, action, reward, agent model, and task. In our taxonomy, we organize existing explainable RL research based on these components:~agent model-explaining methods that show the decision-making mechanism of agent, reward-explaining methods that show how different factors within reward function influence agent policy, state-explaining methods that illustrate the state features at different time stages affecting agent behavior, and task-explaining methods that explain how the agent divide the complex task into subtasks and complete them in long term. These XRL methods are further categorized based on the employed technique. We present this taxonomy in Figure~\ref{Fig1}. The detailed descriptions of these methods are presented in the following subsection.

\subsection{Agent Model-explaining}

Classical RL frameworks primarily aim to optimize the decision-making capability of agent without focusing on the internal decision-making logic. In contrast, agent model-explaining XRL methods achieve high-performing agents while also extract the underlying decision-making mechanism of agent model to generate explanations. We categorize current agent model-explaining XRL methods into two types: self-explainable and explanation-generating techniques. Self-explainable methods aim to generate explanations by the transparent inner agent model itself, whereas explanation-generating techniques provide explanations based on predetermined reasoning mechanisms.
\subsubsection{Self-explainable}

\label{sec::Self-Explainable}
\begin{table}[!t]
% \vspace{-0.2cm}
\caption{Self-explainable agent models in XRL approaches. The venue with $^*$ denotes that the paper published at the workshop of that venue. }
% \vspace{-0.2cm}
\begin{center}
\Large
\setlength{\tabcolsep}{2.2pt}
\resizebox{0.49\textwidth}{!}{
\begin{tabular}{c|c|c|c|c|c}
    \toprule
    \textbf{Type} & \textbf{Explanation} & \textbf{Reference} & \textbf{Year} & \textbf{Venue} & \textbf{Evaluation}\\
    \midrule
    \multirow{2}{*}{Value-based}  & Decision Tree & \cite{LMUT} & 2018 & ECML-PKDD &O.DP, S.D \\\cmidrule{2-6}
    & Formula Expression & \cite{Formula_base} & 2012 & DS &O.DP, S.D \\\midrule

    \multirow{20}{*}{  Policy-based }  & \multirow{4}{*}{Programmatic policy} & \cite{PIRL} & 2018 & ICML &O.DP, S.D \\
    &  & \cite{verma2019imitation} & 2019 & NeurIPS &O.DP, S.D \\
    &  & \cite{inala2020neurosymbolic} & 2020 & NeurIPS &O.DP, S.D \\
    &  & \cite{trivedi2021learning} & 2021 & NeurIPS  &O.DP, S.UP, S.D \\\cmidrule{2-6}
    & \multirow{4}{*}{Symbolic policy} & \cite{Formula_2} & 2018 & EAAI &O.DP, S.D \\
    &  & \cite{Formula_3} & 2019 & GECCO &O.DP, S.D \\
    &  & \cite{landajuela2021discovering} & 2021 & ICML &O.DP, S.D\\
    &  & \cite{delfosse2024interpretable} & 2023 & NeurIPS & O.DP, S.D \\\cmidrule{2-6}
    & \multirow{3}{*}{Fuzzy controller} & \cite{Fuzzy_controller_2} & 2017 & EAAI &O.DP, S.D \\
    &  & \cite{akrour2019towards} & 2019 & ECML-PKDD$^*$ &O.DP, S.D \\
    &  & \cite{ou2023fuzzy} & 2023 & TFS &O.DP, S.D \\\cmidrule{2-6}
    & \multirow{3}{*}{Logic rule} & \cite{NLRL_1} & 2019 & ICML &O.DP, S.D \\
    &  & \cite{NLRL_2} & 2019 & arXiv &O.DP, S.D \\
    &  & \cite{NLRL_3} & 2020 & arXiv &O.DP, S.D \\\cmidrule{2-6}
    & \multirow{7}{*}{Decision Tree} & \cite{DAGGER} & 2011 & ICML &O.DP, S.D \\
    &  & \cite{VIPER} & 2018 & NeurIPS &O.DP, O.R, S.D \\\
    &  & \cite{topin2019generation} & 2019 & AAAI &O.DP, O.F, S.D \\
    &  & \cite{custode2020evolutionary} & 2020& arXiv &O.DP, S.D \\\
    &  & \cite{Conservative_Q_Improvement} & 2020 & arXiv &O.DP, S.D \\\
    &  & \cite{topin2021iterative} & 2021 & AAAI &O.DP, S.D \\\
    &  & \cite{milani2022maviper} & 2022 & arXiv &O.DP, S.D \\\bottomrule
\end{tabular}
}
  \vspace{-0.2cm}

 \end{center}
\label{self-Explainable}
\end{table}%

A self-explainable model is intentionally designed to be self-explanatory throughout the training process, which is accomplished by imposing limitations on the complexity of the model structure~\cite{self-interpretable-1, self-interpretable-2}. Such a model is also known as an intrinsic model~\cite{self-interpretable-2}, as it embodies transparency and ease of understanding. The explanation logic is inherently integrated within the agent model itself.
Our work provides a comprehensive overview of the current self-explainable agent model in XRL field and categorizes them into two types based on the target of explainable agent model: value and policy. This classification can be found in Table~\ref{self-Explainable}.
  % \vspace{-0.2cm}
\paragraph{Value-based}

The Q-value in RL measures the expected discounted sum of rewards that an agent would receive from a given state $(s, a)$. This value can also be employed to construct a deterministic or energy-based policy\cite{SAC,DQN,DDPG}. Thus many value-based agent model-explaining XRL frameworks primarily concentrate on the Q-value model.

The Linear Model U-Tree (LMUT)~\cite{LMUT} combines the concepts of imitation learning (IL) and continuous U-tree (CUT)~\cite{CUT}, which can be considered as an advanced version of CUT for value function estimation.
Similar to a typical decision tree, LMUT internal nodes store dataset features, while the leaf nodes represent a partition of the input space. However, in LMUT, each leaf node contains a linear model that approximates the Q-value instead of a simple constant.
The Q-value approximation, denoted as $Q_{N_t}^{UT}$, is obtained from the linear model within the corresponding LMUT leaf node. This approximation acts as an explanation by quantifying the individual effects of different features in LMUT.
The researchers outline the training process for LMUT, which involves two steps: (1) data gathering phase counting all transitions $T$ within LMUT and modifying Q-values; (2) node splitting phase by Stochastic Gradient Descent~(SGD). When SGD fails to yield sufficient improvement on specific leaf nodes, the framework splits those leaf nodes to disentangle the mixed features.
Experimental results demonstrate that LMUT achieves comparable performance to neural network-based baselines across various environments.

\cite{Formula_base} introduced a search algorithm exploring the space of simple closed-form formulas to construct Q-value.
The variables within the formula represent the abstractions of state and action components, while the operations performed on these variables are unary and binary mathematical operations. The resulting policy is a greedy deterministic policy that selects the action with the maximum Q-value. The different operations highlight the varying effects of variables on the Q-value, thereby ensuring explainability.
However, this method struggles with combinatorial explosion during the search process. Therefore, the total number of variables, constants, and operations is restricted to a small number.
  % \vspace{-0.2cm}
\paragraph{Policy-based}
\begin{figure*}[htbp]
  
	\begin{subfigure}{0.495\textwidth}
		\centering
		\includegraphics[width=\textwidth]{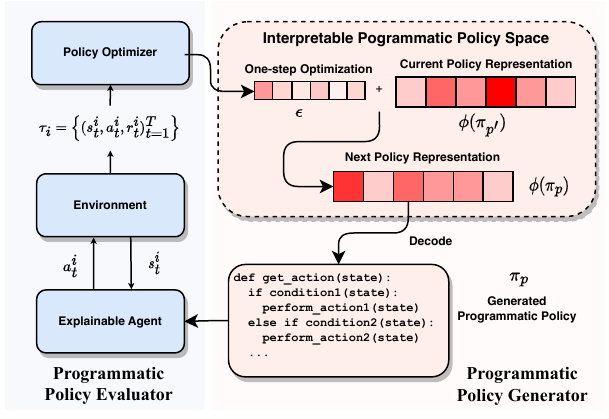}
		\caption{Programmatic policy approaches}
    \label{PRL}
	\end{subfigure}
	\begin{subfigure}{0.495\textwidth}
		\centering
		\includegraphics[width=\textwidth]{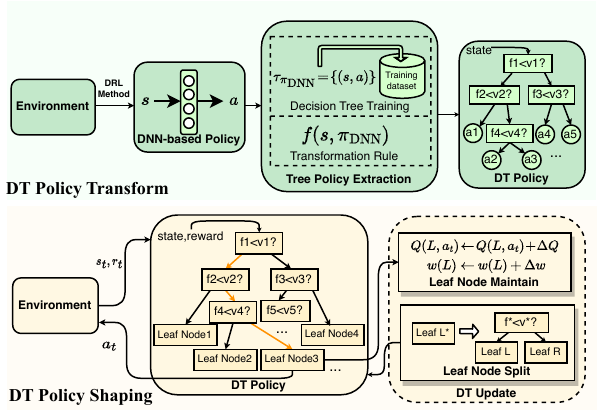}
		\caption{Decision tree policy approaches}
    \label{DT}
	\end{subfigure}
  \caption{
  Examples of Self-Explainable Policy Architectures:~(a) Programmatic reinforcement learning frameworks~\cite{inala2020neurosymbolic,trivedi2021learning,PIRL,verma2019imitation}; (b) Decision tree policy construction by transforming~\cite{VIPER,milani2022maviper} or shaping~\cite{LMUT,Conservative_Q_Improvement}.
  }

  \label{Policy}
\end{figure*}

Policy representation is considered a more direct approach compared to the Q-value, as it explicitly reflects the agent decision-making capability. In this section, we present a comprehensive analysis of potential policy models proposed in the existing literature. Specifically, Figure~\ref{Policy} illustrates several representative approaches of this kind.

Programmatic Reinforcement Learning~(PRL) involves utilizing a program as the representation of the policy, enabling intrinsic explainability through logic rules within the program~\cite{PIRL, verma2019imitation, inala2020neurosymbolic, trivedi2021learning}. This approach operates through two components, as shown in Figure~\ref{PRL}: programmatic policy generator and programmatic policy evaluator. The former updates the current programmatic policy vector within a fixed programmatic space, generating a programmatic policy through vector decoding. The latter involves simulating the generated programmatic policy to optimize the current policy in a one-step fashion. The main challenge in PRL lies in selecting an interpretable programmatic policy space. 
\cite{PIRL} constructs programmatic policy using a domain-specific high-level programming language based on historical data utilization, allowing for a quick understanding of past interactions influence. They propose Neurally Directed Program Search~(NDPS) to construct such policy. NDPS employs DRL method to find a neural policy that approximates the target policy, followed by iterative policy updates through template enumeration using \emph{Bayesian Optimization}~\cite{Bayesian} or satisfiability modulo theory to optimize the parameters. However, \cite{verma2019imitation} argue that this method is highly suboptimal and propose a new framework based on mirror descent-based meta-algorithm for policy search in the space combining neural and programmatic representations. For multi-agent communication, \cite{inala2020neurosymbolic} synthesize programmatic policies based on the generated communication graph of the agents. Additionally, \cite{trivedi2021learning} learn a latent program space to improve the efficiency of programmatic policy search. Furthermore, the learned latent program embedding can be transferred and reused for other tasks.

Formulaic expressions are also able to represent policies instead of value functions. Such policies are referred to as symbolic policies, comprising simple and concise symbolic operations that provide intrinsic explainability through succinct mathematical expressions~\cite{lu2016using,Formula_2,Formula_3}. However, searching the entire symbolic space to find the best fit is generally considered a computationally complex problem known as NP-hard~\cite{lu2016using}. To address this challenge, several studies~\cite{Formula_2,Formula_3} utilize genetic programming for model-based batch RL to maintain a population of symbolic expression individuals as well as evolutionary operations. In contrast to a direct search for a symbolic policy, recent methods~\cite{landajuela2021discovering, delfosse2024interpretable} utilize inexplicable DNN-based anchor policy to generate an explainable symbolic policy.

The policy can be constructed based on the combination of several fuzzy controllers~\cite{akrour2019towards,Fuzzy_controller_2,ou2023fuzzy,Survey_2}.
Specifically, the agent policy, denoted as $\pi(a|s)$, can be represented as a Gaussian Distribution $\mathbf{N}(a|K\varphi(s),\varSigma)$, with $K$ stacking actions for the cluster centers, $\varphi(s)$ returning a weight vector based on the distance, and $\varSigma$ being a state-independent full variance matrix. 
By evaluating the distance to cluster centers, the influence of different centers on actions can be analyzed. \cite{akrour2019towards,ou2023fuzzy} employ policy gradient method to facilitate training of such policies. Additionally, \cite{Fuzzy_controller_2} applies parameter training on a world model to construct fuzzy RL policies. 

First-order logic~(FOL) serves as a foundational language to depict entities and relationships~\cite{barwise1977introduction}. It underpins the policy representation in Neural logic RL~(NLRL), which fuses policy gradient techniques with differentiable inductive logic programming~\cite{zimmer2021differentiable}. The seminal work~\cite{NLRL_1} on NLRL shows the enhancement of explainability through weighted logic rules, clarifying the rationale for action choices. Later advances assign weights to rule atoms, leveraging genetic programming technique for policy formula learning from historical interactions~\cite{NLRL_2,NLRL_3}. This evolution positions NLRL as a tool for deriving potent policies with superior explainability and generalizability.

Decision Tree~(DT) for XRL has been categorized into policy-based and value-based strategies. While the linear model U-tree stands out as a DT variant in value-based XRL, DT-based policies are utilized to select actions based on distinctive features derived from DTs, thereby providing interpretable observations within RL tasks~\cite{VIPER,DAGGER, milani2022maviper, topin2019generation, topin2021iterative}.
Frameworks for policy-based DTs are delineated in Figure~\ref{DT}. With the efficacy of humans in acquiring policies on DDN via DRL, transforming DNN policies to DT policies is a promising strategy. For this, Verifiability via Iterative Policy Extraction~(VIPER)\cite{VIPER} employs model distillation\cite{distillation} to transmute pre-trained DNN policies to DTs using optimal policy trajectories. Techniques like Q-DAGGER~\cite{DAGGER} and MAVIPER~\cite{milani2022maviper} further refine and extend VIPER to more scenarios like multi-agent settings. Iterative Bounding MDP~(IBMDP)~\cite{topin2021iterative} and policy summarization~\cite{topin2019generation} also focus on extracting interpretable policies from DNNs.
Another avenue pursues direct DT policy shaping. By maintaining weight information at the leaf nodes of DT to approximate Q-value and performing leaf node splits at specific stages, a high-performance DT policy can be obtained. \cite{custode2020evolutionary} employ evolutionary algorithms to evolve the DT structure while applying Q-learning to the leaf nodes. \cite{Conservative_Q_Improvement} propose Conservative Q-Improvement~(CQI), which uses lazy updating and expands the tree size only when the approximation of future discount rewards exceeds a specified threshold.

\subsubsection{Explanation-generating}

\begin{table}[!t]
\vspace{0.15cm}
\caption{Explanation-generating agent models in XRL.
% The venue with $^*$ denotes that the paper published at the workshop of that venue.
}
\vspace{-0.15cm}
\large
\begin{center}
\setlength{\tabcolsep}{2.2pt}
\resizebox{0.49\textwidth}{!}{
\begin{tabular}{c|c|c|c|c}
    \toprule
      \textbf{Explanation} & \textbf{Reference} & \textbf{Year} & \textbf{Venue} & \textbf{Evaluation}\\
    \midrule

      \multirow{7}{*}{Counteract} & \cite{AIM}& 2020& AAAI&O.DP, S.UP, S.UC, S.D \\
      & \cite{olson2021counterfactual}& 2021& AI&O.DP, O.F, S.UP, S.UC, S.D \\
      & \cite{stein2021generating} & 2021 & NeurIPS &O.DP, O.F, S.D \\
      & \cite{amitai2022don}& 2022& AAAI&O.DP, S.UC, S.D\\
  & \cite{yu2023explainable} & 2023 & IJCAI &O.DP, S.D \\
  & \cite{meulemans2024would}& 2024& NeurIPS&O.DP, O.F, O.R\\
  &  \cite{amitai2024explaining} & 2024 & AAAI & O.DP, S.UP, S.UC\\\midrule \multirow{2}{*}{Instruction} & \cite{AIM} & 2020 & HAI & O.DP, S.D \\
     & \cite{stein2021generating} & 2021 & ICONIP & O.DP, S.D \\\midrule
     \multirow{3}{*}{Answer to query} & \cite{Query-based} & 2017 & HRI & O.DP, S.D \\
     & \cite{boggess2022toward} & 2022 & IJCAI & O.DP, S.UP, S.UC, S.D \\
     & \cite{boggess2023explainable} & 2023 & IJCAI & O.DP, S.UP, S.UC, S.D \\\midrule
     \multirow{4}{*}{Verify} & \cite{verify} & 2019 & SIGCOMM$^*$ & O.DP, S.D \\
     & \cite{zhu2019inductive} & 2019 & PLDI & O.DP, O.F, S.D \\
     & \cite{anderson2020neurosymbolic} & 2020 & NeurIPS & O.DP, S.D \\
     & \cite{jin2022trainify} & 2022 & CAV & O.DP, O.R, S.D
\\\bottomrule
\end{tabular}
}
 \end{center}
\label{explanation-generating}
\vspace{-0.5cm}
\end{table}%

Explanation-generating methods utilize an explicit auxiliary reasoning mechanism to facilitate the automatic generation of explanations. The development of such a mechanism relies on emulating the human cognitive processes involved in learning novel concepts. Below, we present a selection of influential works that demonstrate these types of explainability, summarized in Table~\ref{explanation-generating}.

Counterfactual explanations answer the question of "why perform X" by explaining "why not perform Y"~(the counterfactual of X)~\cite{olson2021counterfactual,AIM,yu2023explainable,amitai2022don,stein2021generating, amitai2024explaining}. 
\cite{olson2021counterfactual, yu2023explainable,meulemans2024would} crafts counterfactual states $s'$ that have minimal divergence from the current state $s$, yet lead to distinct agent actions, while \cite{stein2021generating} emphasize the Q-value discrepancies between counterfactual action pairs. On the causal front, \cite{AIM} leverages the causal model to grasp the world through distinct variables and potential interrelationships to further elucidate both action reasons and counterfactuals. However, the rigidity of the causal model hampers its adaptability, and the method can only be implemented in discrete action space. To bridge this gap, \cite{yu2023explainable, meulemans2024would} melds attention-driven causal techniques, facilitating causal influence quantification in continuous action spaces and illuminating the long-term repercussions of such actions. Meanwhile, \cite{amitai2024explaining} improves the explainability of counterfactual explanations by visually comparing the actions chosen by the agent with the counterfactual outcomes.

Instruction-based Behavior Explanation~(IBE)~\cite{IBE_1,IBE_2} enhances explainability with formal agent instructions. In basic IBE~\cite{IBE_1}, the agent acquires the capability to explain the behavior with instruction. The learning process includes estimating the target of the agent actions by simulation
and acquiring a mapping from the target of the agent actions to the expressions with a clustering approach. However, it is difficult to divide the state
space to assign an explanation signal in much more complex tasks. Consequently, in their advanced IBE approach~\cite{IBE_2}, DNN model is employed to construct the mapping, enabling its adaptability to intricate state space.

With the pre-defined query template, the agent is able to explain its inner mechanism by answering~\cite{Query-based, boggess2022toward, boggess2023explainable}. \cite{Query-based} introduce a method wherein queries are mapped to decision-making statements via templates. They harness a graph search algorithm to pinpoint relevant states and summarize attributes in natural language. Although the generated policy explanations align with the expert expectations, their reliability in more complex tasks remains unverified. To address this, \cite{boggess2022toward} extend this approach to MARL by proposing Multi-agent MDP~(MMDP), an abstraction of MARL policy. They first transform the learned policy into an MMDP setting by a specified set of feature predicates to address "When, Why not, What" questions in MARL. However, the question templates ignore the task process. To address this drawback, \cite{boggess2023explainable} encodes the temporal query and compares it with the transition model to address temporal queries regarding the task order, resulting in contractive explanations.

Formal verification techniques bolster safety in RL paradigms. Verily~\cite{verify}, for instance, accomplishes verification with the satisfiability modulo theories verification engine for DNN. If the verification result is negative, Verily can generate a counterexample through logical verification to explain the discrepancy. This counterexample can in turn guide the updates of the DNN parameter. \cite{anderson2020neurosymbolic} adopt a similar approach, employing the idea of mirror descent shared by \cite{verma2019imitation}. They perform updating and projecting steps between the neurosymbolic class and restricted symbolic policy class to enable efficient verification. Furthermore, \cite{zhu2019inductive} propose a verification toolchain to ensure the safety of learning DNN-based policies. Likewise, \cite{jin2022trainify} present a verification-in-the-loop framework that iteratively trains and refines the abstracted state space using counterexamples if verification fails.

\subsubsection{Summarization}
Self-explainable methods predominantly utilize S.D as an assessment criterion. These methods leverage the agent model itself to provide explanations and illustrate these explanations through case studies. While case studies showcasing intrinsic explainable policy instances, such as decision tree~\cite{LMUT,topin2019generation} and programmatic policies~\cite{inala2020neurosymbolic, trivedi2021learning}, are intuitively logical and reasonable, the absence of quantitative measurements limits their persuasiveness in demonstrating explainability.
Instead, explanation-generating methods utilize various evaluation assessments such as S.UP, S.UC, O.F, and O.R, which are more formal and quantitative. 
Self-explainable methods primarily rely on agent architecture to provide intrinsic explanations, which are highly formulaic and objectively described in detail. Therefore, objective assessments of O.F and O.R offer a more precise evaluation of the architecture-based explanation. Conversely, explanation-generating methods need more subjective assessments of S.UP and S.UC from human feedback since the core of explanation-generating methods is the extrinsic reasoning mechanisms conducting logical reasoning, which can be effectively evaluated via human participants with strong inferential abilities.

\subsection{Reward-explaining}
\label{sec::re}

Reward-explaining methods reconstruct an explainable reward function through the quantification of various key factors that are instrumental in accomplishing the task, such as the degree of multi-agent cooperation and the features of the final goal.
The reward function plays a crucial role in RL tasks, serving as the primary factor for estimating actions in the short term and policies in the long term. The reward-explaining method involves explicitly designing an explainable reward function to provide explanations for the critical factors of the task.
Building upon this notion, we categorize current reward-based XRL work into two types: reward shaping and reward decomposition. The approaches within each category are summarized in Table~\ref{reward_explaining}.

\subsubsection{Reward Decomposition}
Reward decomposition methods aim to explain the inexplicable value of a reward function by breaking it down into several distinct parts that represent different aspects. The original reward function is a single scalar value influenced by multiple implicit factors. By decomposing the reward function, we can analyze the influence and relationships among these implicit factors. 

Horizontal reward decomposition~\cite{reward-Decomposition} decompose the reward function in the MDP horizontally as $\vec{R}:\mathcal{S}\times\mathcal{A}\rightarrow \mathbb{R}^{|\mathcal{C}|}$, where $\mathcal{C}$ represents the number of reward components. Subsequently, the Q-value is also decomposed as $Q^\pi(s,a)=\sum_{c\in\mathcal{C}}Q^\pi_c(s,a)$. 
To explain the decomposition, the authors primarily focus on comparing pairwise actions. One straightforward approach is Reward Difference eXplanation~(RDX) in the form of $\Delta(s,a_1,a_2)=\vec{Q}(s,a_1)-\vec{Q}(s,a_2)$. RDX informs experts about which components may have an advantage over other factors, but it does not identify the most significant component. Moreover, RDX may offer limited explanations when the number of factors increases. To address this, the authors propose another form of explanation called Minimal Sufficient eXplanation~(MSX). MSX is a two-tuple $({\rm MSX}^+,{\rm MSX}^-)$, with ${\rm MSX}^+$ selecting the minimal set of components where the total $\Delta(s,a_1,a_2)$ surpasses a dynamic threshold, while ${\rm MSX}^-$ checks the summation of $-\Delta(s,a_1,a_2)$ with the other threshold. Similarly, in vision-based RL, \cite{liu2023visual} reconstruct the multidimensional patch reward of image samples by assessing the expertise of each local patch. The patch reward effectively captures features, serving as a fine-grained measure of expertise and a tool for visual explainability.

\begin{table}[!t]
\vspace{0.15cm}
\caption{Reward-explaining methods in XRL.}
\vspace{-0.2cm}
\begin{center}
\setlength{\tabcolsep}{2.2pt}
\resizebox{0.49\textwidth}{!}{
\begin{tabular}{c|c|c|c|c}
    \toprule
     \textbf{Type} & \textbf{Reference} & \textbf{Year} & \textbf{Venue} & \textbf{Evaluation}\\
    \midrule

     \multirow{5}{*}{Reward decomposition}  & \cite{foerster2018counterfactual} & 2018 & AAAI &O.DP, S.D \\
    & \cite{reward-Decomposition} & 2019 & IJCAI$^*$ &O.DP, S.D \\
    & \cite{shapley_Q-value} & 2020 & AAAI &O.DP, S.D \\
    & \cite{li2021shapley} & 2021 & SIGKDD &O.DP, S.D \\
    & \cite{liu2023visual} & 2023 & ICLR &O.DP, S.D \\\midrule
    
    \multirow{7}{*}{Reward shaping}  & \cite{RARE} & 2019 & HRI & O.DP, S.D \\
    & \cite{reward-level} & 2019 & AAAI & O.DP, S.D \\
    & \cite{wu2020tree} & 2020 & AAAI & O.DP, S.D \\
    & \cite{wu2021self} & 2021 & AAAI & O.DP, S.D \\
    & \cite{mirchandani2021ella} & 2021 & NeurIPS & O.DP, S.D \\
    & \cite{jin2022creativity} & 2022 & AAAI &O.DP, S.D\\
    & \cite{ashwood2022dynamic} & 2022 & NeurIPS & O.DP, S.D
\\\bottomrule
\end{tabular}
}
 \end{center}
\label{reward_explaining}
\vspace{-0.5cm}
\end{table}%

For multi-agent tasks, the widely adopted paradigm is Centralized Training with Decentralized Execution~(CTDE), which allows agents to train based on their local view while a central critic estimates the joint value function. The primary challenge of CTDE lies in assigning credit to each agent. One effective tool for assigning credit to each local agent is the Shapley value~\cite{Shapley_base}, which represents the average contribution of an entity~(or in the context of multi-agent RL, a single agent) across different scenarios. To compute the Shapley value, we can measure the change in the output when the target feature or agent is considered. Considering that the computational costs growing exponentially with the number of agents make it hard to approximate in complex environments, \cite{foerster2018counterfactual} employs a counterfactual advantage function for local agent training. Nevertheless, this method neglects the correlation and interaction between local agents, leading to failure on intricate tasks. To address this limitation, \cite{shapley_Q-value} combine the Shapley value with the Q-value and perform reward decomposition at a higher level in multi-agent tasks to plan global rewards rationally: individual agents with greater contributions receive more rewards. Therefore, this method assigns credit to each agent, enabling an explanation of how the global reward is divided during training and how much each agent contributes. However, this network-based method relies highly on the assumption that local agents take actions sequentially without considering the synchronous running of agents. In contrast, \cite{li2021shapley} employ counterfactual-based methods to quantify the contribution of each agent, which proves to be more stable.

\subsubsection{Reward Shaping}
Direct synthesis of explainable reward functions provides an alternative pathway, as demonstrated by notable research efforts~\cite{RARE,reward-level,jin2022creativity,wu2020tree,wu2021self,mirchandani2021ella}. These approaches establish reward representations through explicit structural alignment with task objectives, eliminating the need to explain the pre-existing reward function components.

% Directly obtaining an explainable reward function is also a viable approach. Several studies~\cite{RARE,reward-level,jin2022creativity,wu2020tree,wu2021self,mirchandani2021ella} have taken the route of directly seeking understandable reward function that explicitly captures the task structure, bypassing the need to explicitly explain the reward function component.

Building upon the interactions between the agent and humans, \cite{mirchandani2021ella} present a reward-shaping approach that modifies the original sparse rewards with human instruction goals into a dense explainable rewards. Similarly, The study conducted by \cite{ashwood2022dynamic} employs a meta-learning approach to acquire multiple goal maps, subsequently modifying the reward function by aggregating diverse goal map weights. \cite{RARE} propose a framework that employs the Partially Observable MDP~(POMDP) model to approximate collaborator understanding of joint tasks. The authors continually modify and correct the reward function in order to achieve this objective. If they discover a more plausible reward function, they evaluate whether the advantage of adopting it outweighs the cost of abandoning the previous function. Subsequently, a repairing representation is generated if the newly found reward function proves beneficial. 

To enhance the explainability of complex tasks, the adoption of multi-level rewards is a viable approach. Unlike task decomposition, where decomposed reward reflects the actual rewards from the environment, multi-level reward encompasses both extrinsic rewards from the environment and intrinsic rewards aimed at facilitating comprehension and explanation. \cite{reward-level} introduce a two-level framework comprising extrinsic reward standing for real rewards within RL environments and intrinsic reward representing the achievement of inner task factors. Symbolic Planning approaches are utilized to maximize the intrinsic reward. Meanwhile, compared to \cite{reward-level} utilizing the predefined intrinsic reward to generate plans, \cite{jin2022creativity} extend their work by automatically learning the intrinsic reward, enabling faster convergence compared to the original approach. For the task of temporal language bounding in untrimmed videos, \cite{wu2020tree} propose a tree-structured progressive RL technique: while the leaf policy receives the extrinsic reward from the external environment, the root policy, which does not directly interact with the environment, evaluates rewards intrinsically based on high-level semantic branch selections.  
Meanwhile, to address challenges with defining intrinsic rewards resulting in inferior performance compared to extrinsic rewards, \cite{wu2021self} introduce the concept of intrinsic mega-rewards to enhance the agent individual control abilities, including direct and latent control. A relational transition model is formulated to enable the acquisition of such control abilities, yielding superior performance compared to existing intrinsic reward approaches.

\subsubsection{Evaluation}
% The reward-explaining method entails analyzing various factors in the task that profoundly influence agent behavior and incorporating them into the reward function to obtain RL explainability,which quantifies the influence of relevant task aspects to elucidate agent decisions within the updated reward function, thereby providing a clear and detailed depiction of the different task factor influence~\cite{reward-Decomposition}. Moreover, Considering that the reward function in the MDP context is typically crafted manually with significant effort of RL researchers to improve agent performance, we posit that such explainable reward functions, which offer insights into how rewards affect the agent, can inform and guide the reverse design of reward function for RL researchers. However, it should be acknowledged that comprehending the rewards in MDP framework may present challenges for people lacking a background in RL, which makes it challenging to verify such explanation via human users directly.}

All of the surveyed reward-explaining methods utilize S.D to assess RL explainability. These explanations in the original paper case study provide visualizations comparing the influences of different aspects of the task within the reward value, offering illustrative examples for case studies in the paper. In order to perform quantitative analysis of these reward-explaining methods in the future, we posit that objective assessment O.F is more required for future research. Different quantitative influences of factors are the backbone of reward explanation, which should be strictly measured for its fidelity objectively. A possible way to measure O.F is manually updating the reward against the reward explanation to see whether the performance falls rapidly~\cite{guo2021edge}.

\subsection{State-explaining}
\label{sec::state_explaining}
\qyp{State-explaining methods generate extrinsic explanations based on observation from the environment, which incorporate a state analyzer that allows for the simultaneous analysis of the different state features significance. According to the time stage of different states to construct the explanation, we divide current state-explaining methods into three types: historical trajectory-based methods focusing on past significant states, current observation-based methods emphasizing important features of current state, and future prediction-based methods inferring future states. We provide a brief review of the relevant literature on state-level explainability in Table~\ref{state-explaining}.}
\begin{table}[!t]
\vspace{0.15cm}
\caption{State-explaining methods in XRL.}
\vspace{-0.2cm}
\Large
\begin{center}
\setlength{\tabcolsep}{2.2pt}
\resizebox{0,49\textwidth}{!}{
\begin{tabular}{c|c|c|c|c}
    \toprule
     \textbf{Temporal perspective} & \textbf{Reference} & \textbf{Year} & \textbf{Venue} & \textbf{Evaluation}\\
    \midrule

     \multirow{10}{*}{Historical trajectory}  & \cite{SBRL} & 2014 & AAAI &O.DP, O.R, S.D \\
    & \cite{V-SBRL} & 2018 & SSCI &O.DP, S.D \\
    & \cite{introspection} & 2018 & AI &O.DP, S.UP, S.UC, S.D \\
    & \cite{shapley_his} & 2021 & TCSS &O.DP, S.D \\
    & \cite{guo2021edge} & 2021 & NeurIPS &O.DP, O.F, O.R, S.D \\
    & \cite{heuillet2022collective} & 2021 & CIM &O.DP, S.D \\
    & \cite{kenny2022towards} & 2022 & NeurIPS &O.DP, O.F, O.R, S.D \\
    & \cite{ragodos2022protox} & 2023 & ICLR &O.DP, S.UP, S.UC, S.D \\
    & \cite{deshmukh2023explaining} & 2023 & ICLR &O.DP, O.R, S.D \\
    &  \cite{sun2024accountability} & 2023 & NeurIPS & O.DP, S.UP \\ \midrule
    \multirow{19}{*}{Current observation}  & \cite{LMUT} & 2018 & ECML-PKDD & O.DP, S.D \\
    & \cite{RS-rainbow} & 2018 & arXiv & O.DP, S.D \\
    & \cite{goel2018unsupervised} & 2018 & NeurIPS & O.DP, S.D \\
    & \cite{perturbation-based-saliency} & 2018 & ICML & O.DP, O.R, S.D \\
    & \cite{petsiuk2018rise} & 2018 & ICML & O.DP, S.D \\
    & \cite{object-saliency-map} & 2018 & AIES & O.DP, S.UP \\
    & \cite{leurent2019social} & 2019 & arXiv & O.DP, S.D \\
    & \cite{DQNViz} & 2019 & TVCG & O.DP, S.D \\
    & \cite{annasamy2019towards} & 2019 & AAAI & O.DP, O.R, S.D \\
    & \cite{neuroevolution} & 2020 & GECCO & O.DP, O.R, S.D \\
    & \cite{xu2020deep} & 2020 & NeurIPS & O.DP, S.D \\
    & \cite{pan2020xgail} & 2020 & KDD & O.DP, S.D \\
    & \cite{tang2021sensory} & 2021 & NeurIPS & O.DP, O.R, S.D \\
    & \cite{guo2021machine} & 2021 & NeurIPS & O.DP, S.UP, S.UC, S.D \\
    & \cite{waldchen2022training} & 2022 & ICML & O.DP, O.F, S.D \\
    & \cite{bertoin2022look} & 2022 & NeurIPS &O.DP, S.D \\
    & \cite{peng2022inherently} & 2022 & NeurIPS &O.DP, S.UP, S.UC, S.D \\
    & \cite{beechey2023explaining} & 2023 & ICML &O.DP, S.D \\
 & \cite{wang2024explainable}& 2024& TIV &O.DP, O.R, S.D\\\midrule
    \multirow{6}{*}{Future prediction}  & \cite{van2018contrastive} & 2018 & arXiv &O.DP, O.R, S.D \\
    & \cite{SPC} & 2019 & ICRA &O.DP, S.D \\
    & \cite{Monte-Carlo-Dropout} & 2019 & ICRA &O.DP, S.D\\
    & \cite{yau2020did} & 2020 & NeurIPS &O.DP, S.D \\
    & \cite{lee2020weakly} & 2020 & NeurIPS &O.DP, S.D\\
    &  \cite{hu2023explainable} & 2023 & TTE & O.DP, S.D\\
    &  \cite{lee2024refining} & 2023 & NeurIPS & O.DP, S.D\\
\bottomrule
\end{tabular}
} 

 \end{center}
\label{state-explaining}
\vspace{-0.6cm}
\end{table}%

\subsubsection{Historical Trajectory}

Starting from the trace of historical decisions, numerous studies aim to estimate the influence of historical observations on future decision-making by agents.

Sparse Bayesian Reinforcement Learning~(SBRL)~\cite{SBRL} constructs latent space representing past experiences during training to facilitate knowledge transfer and continuous action search. SBRL offers an intuitive explanation for how historical data samples impact the learning process. Another approach, Visual SBRL~(V-SBRL)~\cite{V-SBRL}, utilizes a sparse filter to maintain the significant past image-based state while discarding the trivial ones, resulting in a sparse image set containing valuable past experience. \cite{introspection} identify interestingness state elements in historical observations from various aspects, such as the reward outliers and environment dynamics, and present the detected observations in video. \cite{ragodos2022protox, deshmukh2023explaining, sun2024accountability} denote the important past experiences as prototypes, which are learned by contrastive learning during training. By integrating these prototypes into the policy network, human users are able to observe representative interactions within the task. Meanwhile, \cite{ragodos2022protox} generate a broader explanation by comparing current policy output with human-defined prototypes, demonstrating better trustworthiness and performance.

The Shapley value also offers an effective approach for calculating and visualizing the contribution of each feature in prior trajectories. However, the naive computation of the Shapley value faces an exponential complexity. To mitigate this issue, \cite{heuillet2022collective} employ Monte Carlo sampling to approximate the Shapley value, while \cite{shapley_his} leverage DNN to compute the feature gradients and aggregate them as a Shapley value to develop a 3D feature-time-SHAP map to visualize the significance of each timestep.

Previous surveyed methods only focus on the historical interactions within an episode, \cite{guo2021edge} extend their horizon by considering interactions across episodes. They incorporate a deep recurrent kernel of the Gaussian Process that takes inputs of timestep embeddings capture the correlation between timesteps as well as the cumulative impact across episodes. Furthermore, these outputs can be employed for episode-level reward prediction via linear regression analysis. The regression coefficients obtained from the linear regression model can identify important timesteps, thereby enhancing the explainability of the results.

\subsubsection{Current observation}
 \begin{figure*}[t]
  
	\begin{subfigure}{0.495\textwidth}
		\centering
		\includegraphics[width=\textwidth]{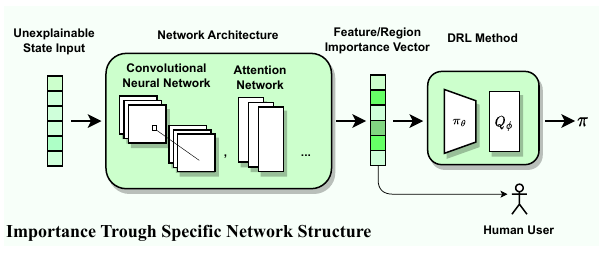}
		\caption{structure-based importance}
    \label{se1}
	\end{subfigure}
	\begin{subfigure}{0.495\textwidth}
		\centering
		\includegraphics[width=\textwidth]{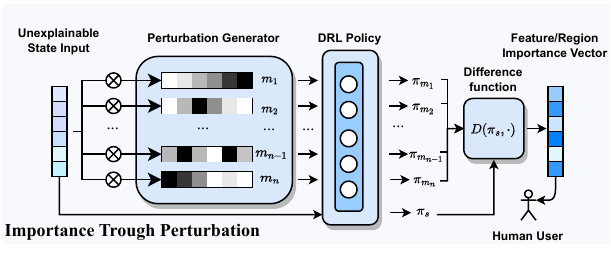}
		\caption{Perturbation-based importance}
    \label{se2}
	\end{subfigure}
  % \vspace{-0.3cm}
  \caption{Examples of state importance extraction techniques via (a) intrinsic architectures~\cite{leurent2019social,tang2021sensory,annasamy2019towards,neuroevolution, peng2022inherently} and (b) extrinsic pertubations~\cite{RS-rainbow,petsiuk2018rise,perturbation-based-saliency,object-saliency-map, bertoin2022look}.}

  \vspace{-0.3cm}
  \label{se}
\end{figure*}
Numerous studies aim to identify critical features influencing decision-making in the current state, particularly in image-based environments. These approaches offer extrinsic explanations by analyzing the impact of state features on agent behavior. Different methods that fall under this category are depicted in Figure~\ref{se}.

The Linear Model U-Tree (LMUT)~\cite{LMUT} evaluates the importance of an LMUT node and its features based on the certainty of the Q-value and the squared weight of the features. The paper applies LMUT to video games and gets pixels with relatively high influence. The explanation denotes such pixels as ``super-pixels'', which are crucial for decision-making.

Several studies leverage self-attention, allowing the creation of an attention score matrix, which highlights relationships among input features for improved explainability~\cite{neuroevolution,RS-rainbow,object-saliency-map,perturbation-based-saliency,leurent2019social,tang2021sensory,annasamy2019towards,xu2020deep, peng2022inherently}. In contexts like agent interaction in self-driving scenarios, self-attention-based DNN~\cite{leurent2019social} discerns relations amongst multiple entities. Extending this, \cite{tang2021sensory} use attention neurons for honing in on specific state components. Meanwhile, \cite{annasamy2019towards} integrate attention within DNNs to develop auto-encoders for input state reconstruction. Neuroevolution combined with self-attention~\cite{neuroevolution} selects spatial patches over individual pixels, enabling the agent to focus on task-critical areas, thus amplifying efficacy and clarity. \cite{RS-rainbow} introduce a region-sensitive module post-DNN to pinpoint essential input image regions, serving as an explainable plug-in module integrated into classical RL algorithms~\cite{A3C,PPO}. Shifting from pixel-centric states, \cite{xu2020deep} design a hierarchical attention model for text-based games using knowledge graph, capturing state feature relationships. Building on this, \cite{peng2022inherently} enhance explanations by integrating multiple subgraphs with template-filling techniques.

Saliency maps, distinguishing from attention by highlighting specific parts of scenes, like objects or regions, have been adopted to increase explainability in RL agents. These maps showcase pixel influences on outputs through gradient measurements of normalized scores. Several notable studies have contributed to XRL domain~\cite{perturbation-based-saliency,object-saliency-map,petsiuk2018rise,wang2019alphastock,guo2021machine,pan2020xgail,waldchen2022training, beechey2023explaining, wang2024explainable}. \cite{petsiuk2018rise} gauged pixel significance by applying a random value mask and evaluating its decision impact, extended by \cite{pan2020xgail} for geographic areas. \cite{perturbation-based-saliency} introduced perturbation-based saliency, perturbing certain features certainties to discern their impacts on policy. This is further employed by \cite{guo2021machine, wang2024explainable} to juxtapose human and RL agent attention patterns, indicating RL training potential to humanize agents. Improving on this, \cite{bertoin2022look} harnessed unsupervised learning for perturbation-based saliency maps and agent training regularization. Meanwhile, \cite{waldchen2022training} applied CNN for partial feature interpretations, whereas \cite{beechey2023explaining} leveraged shapley values to analyze the effects of feature removals. Lastly, the object saliency map~\cite{object-saliency-map} integrates template matching and enables easier human interpretations by connecting pixel saliency maps with object detection.

In contrast to relying on local spatial information, \cite{goel2018unsupervised} utilize flow information to capture and segment the moving object in the image. Therefore, the policy can focus on moving objects in a more interpretable manner. Furthermore, \cite{DQNViz} propose a specialized framework for visualizing DQN~\cite{DQN} process. This visualization provides insights into the operations performed at each stage and the activation levels of each layer within the deep neural network.

\subsubsection{Future Prediction}
The future prediction method first generates forecasts for future events and then uses these predictions to produce various explanations.

A common approach to predicting the future involves repeated forward simulations from the current state~\cite{van2018contrastive}. However, these simulations may be unprecise due to stochastic environmental factors and approximate biases in training~\cite{hasselt2010double}. To address this, \cite{yau2020did} maintains the discounted expected future state visitations with temporal difference loss to further construct the belief map. The training process of such a belief map is consistent with current value-based inexplainable RL frameworks. This advantage renders it an explainable plug-in for value-based RL methods. Meanwhile, \cite{lee2024refining} utilize diffusion model to dynamically generate future state sequences conditioned on current states. \cite{lee2020weakly, hu2023explainable} combine future prediction with multi-goal RL, facilitating trustworthy predictions of goal for the current state. Semantic Predictive Control~(SPC)~\cite{SPC} dynamically learns the environment and aggregates multi-scale feature maps to predict future semantic events. Additionally, \cite{Monte-Carlo-Dropout} employ an ensemble of LSTM networks trained using Monte Carlo Dropout and bootstrapping to estimate the probability of future events and predict uncertainty in new observations.  Recently, 

\subsubsection{Evaluation}
Existing state-explaining methods undergo evaluation through various assessments, encompassing both subjective and objective measures. O.R stands out as the predominantly employed quantitative assessment~\cite{tang2021sensory, annasamy2019towards}. To further evaluate the quality of explanations in various aspects, it is necessary to utilize more objective methods of O.F and O.R to assess the accuracy and robustness of the allocation of importance on both temporal and spatial features. However, subjective measurement is not applicable since gathering a significant amount of evaluation data through human feedback would be time-consuming and ineffective.

\subsection{Task-explaining}
\label{sec::ht}

Task-explaining method explains how to divide the current complex task into multiple subtasks via the hierarchical agent. In a hierarchical agent, a high-level controller selects options, while several low-level controllers choose primitive actions. The option chosen by the high-level controller acts as a sub-goal for the low-level controllers to accomplish.  This division of labor in Hierarchical Reinforcement Learning (HRL) enhances architectural explainability compared to the aforementioned XRL works, offering insight into how the high-level agent schedules the low-level tasks. In this context, we delve into HRL and categorize its approaches into two parts: the whole top-down structure and simple task decomposition according to the scheduling mechanism of high-level agents. These categorized approaches are presented in Table~\ref{task_explaining}.

\begin{table}[!t]
\vspace{0.15cm}
\caption{Task-explaining methods in XRL.}
\vspace{-0.2cm}
\begin{center}
\setlength{\tabcolsep}{2.2pt}
\resizebox{0.49\textwidth}{!}{
\begin{tabular}{c|c|c|c|c}
    \toprule
     \textbf{Type} & \textbf{Reference} & \textbf{Year} & \textbf{Venue} & \textbf{Evaluation}\\
    \midrule

     \multirow{2}{*}{Whole Top-to-Down structure}  & \cite{STG} & 2018 & ICLR &O.DP, S.D \\
    & \cite{bool_task} & 2020 & NeurIPS &O.DP, S.D \\\midrule
    \multirow{5}{*}{Simple task division}  & \cite{jiang2019language} & 2019 & NeurIPS & O.DP, S.D \\
    & \cite{D2D} & 2019 & IROS &O.DP, S.D \\
    & \cite{model_primitives} & 2020 & AAMAS &O.DP, S.D \\
    & \cite{sodhani2021multi} & 2021 & ICML & O.DP, S.D \\
    & \cite{reward-level} & 2021 & TCSS & O.DP, S.D
\\\bottomrule
\end{tabular}
}

 \end{center}
\label{task_explaining}
\vspace{-0.4cm}
\end{table}%
      
\subsubsection{Whole Top-to-Down Structure}
In hierarchical tasks with this structure, task sets are divided into multiple levels. The low-level task sets are subsets of the high-level task sets, with the latter containing task elements absent in the former. This well-defined and coherent structure enhances explainability, as it aligns with human experiences and allows for the observation of how the high-level agent schedules low-level tasks.

A notable study~\cite{STG} train a hierarchical policy in Minecraft , an open-world and multi-task environment. The task division sets, denoted as $G_1,G_2,...,G_k$, follow a hierarchical structure: $G_1\subset G_2\subset ...\subset G_k$. At each level, a policy $\pi_k$ comprises four components: a base task set policy $\pi_{k-1}$, an instruction policy $\pi_k^{inst}$ for providing instructions $g$ to guide the execution of base tasks by $\pi_{k-1}$, an augment flat policy $\pi_k^{Aug}$ that directly selects actions for $\pi_k$ instead of relying on base tasks, and a switch policy $\pi_k^{sw}$ that determines whether to choose actions from the base tasks or the augment flat. The state is represented as the pair $(e_t,g_t)$, where $e_t$ signifies time and $g_t$ represents the instruction. 
To train such a hierarchical policy, a two-step approach is proposed. Firstly, basic skills are learned from $G_{k-1}$ to ensure that the previously acquired policy can be leveraged by instructing the base policy. This stage establishes the connection between the instruction policy and the base policy. Next, samples are collected from $G_k$ to learn new skills and the switch policy. Both steps rely on the classical actor-critic RL algorithm.

Another idea is about the logical combination of base tasks utilizing bool algebra form~\cite{bool_task}. This allows for task expressions to employ logical operations such as disjunction, conjunction, and negation. The proposed framework focuses on lifelong learning, which necessitates the utilization of previously acquired skills to solve new tasks. Consequently, the tasks $G_i$ follow a sequential relation: $G_1\subset G_2...\subset G_{t-1}\subset G_t$. In this framework, the paper initially learns goal-oriented approximations of the value function for each base task and subsequently combines these approximations in a specific manner. By leveraging this framework, it becomes possible to acquire new task skills without the need for additional learning. Additionally, it successfully represents the optimal policy for the current RL task using Boolean algebra.
\subsubsection{Simple Task Division}
In contrast to a strictly top-down structure, where sub-tasks are hierarchically defined, simply divided sub-tasks exhibit equal status and filter out priority over each other. Within the context of multi-task reinforcement learning, an efficient approach is needed for knowledge transfer among tasks. To address this, metadata can serve as a valuable tool for capturing task structures and facilitating knowledge transfer. In a study by \cite{sodhani2021multi}, the authors leverage metadata to learn explainable contextual representations across a family of tasks. These sub-tasks align with a higher-level overarching goal, leading to the division of tasks into two levels. The low-level tasks typically represent decomposed sub-tasks of the original task, sharing the same status. Conversely, the high-level task focuses on scheduling the sub-tasks within the overall task structure.

Numerous methods involve explicitly dividing tasks and constructing a high-level agent as a scheduler for low-level agents. \cite{jiang2019language} train the high-level agent to produce language instructions for the low-level agents. During training, the low-level agents employ a condition-RL algorithm, while the high-level agents use a language model-based RL algorithm. All language instructions generated by the high-level agents are comprehensible to humans. 
The symbolic planning+RL method~\cite{reward-level} employs a planner-controller-meta-controller framework to address hierarchical tasks. The planner operates at a higher level, leveraging symbolic knowledge to schedule sub-task sequences. Meanwhile, the controller operates at a lower level, employing traditional DRL methods to solve sub-tasks, and the meta-controller simultaneously providing a new intrinsic target for the planner to guide better task-solving explicitly. 
In the Dot-to-Dot (D2D) framework~\cite{D2D}, the high-level agent constructs the environment dynamics, and utilized it to provides direction to the low-level agents. The low-level agent receives guidance from the high-level agent and solves decomposed, simpler sub-tasks. As a result, the high-level agent can learn an explainable representation of the decision-making process, while the low-level agent effectively learns the larger state and action space.

Unlike the two aforementioned approaches, \cite{model_primitives} adopt a different strategy for task division. They utilize a primitive model instead of directly dividing the task. Initially, the primitive model approximates piecewise functional decomposition. Each specialized primitive model focuses on a distinct region, resulting in corresponding sub-policies specialized in those regions. The sub-policies are subsequently transferred to compose the complete policy for the desired tasks. Through the combination of these sub-policies, this framework retains architectural explainability. The efficacy of this explainability is demonstrated on high-dimensional continuous tasks, both in lifelong learning scenarios and single-task learning. However, the use of the primitive model may not be individually effective for learning to decompose mixed tasks.

\subsubsection{Evaluation}
Currently, all task-explaining methods employ S.D for evaluation. Task-explaining methods provide explanation on task decomposition and scheduling. To measure whether the high-level agent scheduling is correct and reasonable, experienced human is able to provide precise criterion. Hence, we propose that subjective assessments of S.UP and S.UC, utilizing the divide-and-conquer approach of human participants as evaluating criteria, are more suitable for judging the effectiveness of task-explaining methods in dividing tasks.

\begin{table}[!t]
\vspace{0.15cm}
\caption{Comparison of different types of  XRL approaches. ``H'' and ``L'' denotes ``High'' and ``Low'' respectively.}
\vspace{-0.15cm}
\Huge
\begin{center}
\setlength{\tabcolsep}{2.2pt}
\resizebox{0.49\textwidth}{!}{
\begin{tabular}{c|c|c|c|c|c}
    \toprule
     \textbf{Type} & \textbf{Quantification} & \textbf{Fineness} & \textbf{verifiability} &\textbf{Clarity} &\textbf{Need RL prior} \\
    \midrule
Agent model-explaining & H & H & H & L & H\\
Reward-explaining & H & H & L & H & H\\ 
State-explaining & H & H & L & H & L \\  
Task-explaining & L & L & H & H & L\\\bottomrule
\end{tabular}
}
 \end{center}
\label{table::compare}
\vspace{-0.4cm}
\end{table}%

\subsection{XRL Methodology Selection}
Despite comprehensive documentation of our taxonomic framework and extensive literature analysis, persistent uncertainties remain regarding selection criteria among XRL methodologies. To address this gap, we present a systematic comparative analysis of XRL archetypes in 
Table~\ref{table::compare}, derived through summarization across the four methodological categories. This analytical framework enables practitioners to align methodological selection with both system requirements and explainable objectives through feature-based evaluation.

Considering the beginner for a specific task, the intrinsic RL explanation helps the beginner to understand the task and its solving process. Individuals unfamiliar with a particular task can quickly grasp its structure, objectives, and general problem-solving approaches through task-explaining methods~\cite{STG, model_primitives}. To gain more specific and professional insights for solving the task comprehensively, the agent model-explaining methods can guide human learning by providing a policy sketch that illuminates the internal reasoning of XRL agents~\cite{PIRL, VIPER, NLRL_1, verma2019imitation}. Experienced human users who can only solve the task suboptimally possess a general understanding of the method for solving the task but lack proficiency in the specific behaviors required. Therefore, capturing crucial observational features from state-explaining methods can significantly enhance their short-term decision-making. Simultaneously, the reward-explaining methods highlight various latent aspects in the task that contribute to performance changes, enabling human users to attentively select actions and improve their long-term performance.

For RL researchers, different types of XRL methods help the RL researchers to get insight into the exploration of agents and dynamics of environments. Agent model-explaining methods~\cite{zhong2022improving, AIM} illustrate how the inner mechanism of the agent changes during training. And task-explaining methods~\cite{mirchandani2021ella, ashwood2022dynamic} provide valuable insights regarding the task complexity. These insights can be leveraged to guide fine-tuning of agent architecture for improving performance. Meanwhile, by receiving the explanation of how different factors in the reward function affect the policy of agent~\cite{shapley_his, shapley_Q-value}, researchers can gain a better understanding of how to design an effective reward function to enhance the agent performance~\cite{liu2023curricular, luss2023local}. State-explaining methods shed light on the dynamic focus of the agent on state features during training, enabling researchers to comprehend how these features affect the agent's decision-making process. 

Different human groups can choose different XRL methods based on their specific requirements to augment their comprehension of tasks and successfully complete them.

\section{Human knowledge for XRL}
\label{sec::Human_knowledge_for_RL_paradigm}
Incorporating human prior knowledge into XRL enhances agent performance and explainability by aligning learning objectives with domain expertise. While mainstream XRL frameworks often neglect human participation during training, emerging studies demonstrate its benefits in guiding agent behavior and explanatory quality~\cite{KoGuN,LEARN,GAZE1,GAZE2,guan2021widening,TASK}. 
Given their underrepresentation in current XRL research, we advocate for systematic integration of human knowledge within existing taxonomic frameworks to advance XRL.

% Human prior knowledge-integrated XRL methods have shown high performance and explainability.
% Currently, the mainstream XRL frameworks previously discussed in our categorization overlook the potential influence of human participation during training. Nonetheless, several studies have demonstrated the advantages of human participation in XRL~\cite{KoGuN,LEARN,GAZE1,GAZE2, guan2021widening,TASK}. To underscore the significance of human knowledge-intergrated approaches and promote future research in this area, we dedicate a separate section for their discussion. Human participation helps to incorporate prior human knowledge about the tasks as evaluation criteria and guidance, facilitating high-quality agent behaviors and corresponding explanations, compared to classical RL and normal XRL training methods~\cite{rajendran2022human,menda2019ensembledagger,ramakrishnan2019overcoming}. Although the human knowledge may not align perfectly with the task, the agent endeavors to effectively utilize and implement it during the training process. This optimization process aligns with the natural human learning process, wherein expert guidance and knowledge are often imprecise but still valuable for efficient learning. Given the efficiency of human knowledge-integrated XRL and its limited presence in the current XRL community, we emphasize the importance of this approach by presenting existing works organized according to our taxonomy of XRL.

\subsection{Fuzzy Controller Representing Human Knowledge}
Fuzzy logic can be utilized to represent human knowledge to further construct the agent policy, obtaining intrinsic RL explainability on agent architecture.
% As discussed in agent model-explaining approaches within XRL, self-explainable agent models are leveraged to approximate the Q-value or policy in the RL framework. However, a crucial consideration in incorporating human knowledge is how to bridge the comprehension gap between human and RL agent.	 Human knowledge frequently exhibits imprecision and vagueness, particularly when applied to novel tasks, where it may only pertain to a limited subset of the state space.
Traditional approaches, such as bivalent logic rules, are ill-suited for the representation of vague human knowledge due to the overly deterministic nature. In contrast, fuzzy logic can effectively represent human knowledge in an uncertain and imprecise manner.
A notable contribution in this area is the work of \cite{KoGuN}, who propose the KoGuN policy network utilizing the knowledge controller to integrate human suboptimal knowledge. The knowledge controller utilizes a set of fuzzy rules $\left\{l_i\right\}$ translated from human knowledge. Given the state input $s$, these fuzzy rules representing prior human knowledge jointly output a preference action $\mathbf p$., which is then fine-tuned by summing with an additional vector $\mathbf p'$ produced by a hypernetwork. Meanwhile, to tackle the challenge of possible human knowledge mismatch under different states, trainable rule weights $\beta_i$ are introduced for each rule $l_i$ in order to facilitate adaptation to new tasks and optimize the performance of the knowledge controllers. This policy network, with prior human knowledge, is trained by the conventional PPO method. Although the final action output is slightly fine-tuned by the hypernetwork, the rules weights $\left\{\beta_i\right\}$ effectively illustrate the influence of different human knowledge towards agent decision-making, exhibiting the intrinsic explainability of agent architecture. Similarly, \cite{rudolf2022fuzzy} incorporate exact traffic laws into fuzzy logic rules to constraint the self-vehicle behaviors, and \cite{shi2022efficient} establish fuzzy rules containing human knowledge into hierarchical policy, enhancing both fast training and explainability.
\subsection{Dense Reward on Human Command}
The intrinsic dense reward can not only release the ineffective learning under the sparse reward setting but also align with the human command to provide explanation to indicate the agent motivation.
Although sparse reward is frequently employed in real tasks due to its simplicity, learning with sparse reward is challenging. Therefore, efforts have been made to define a dense reward function that provides a reward signal for each action performed. Several studies~\cite{ng1999policy, burda2018exploration, badia2020agent57, harutyunyan2019hindsight, liu2019sequence} have introduced dense reward functions that focus on state-based novelty. Yet, these studies often fail to provide a clear explanation of the underlying motivation and logical sequence of the task goal. An innovative approach presented by \cite{LEARN} introduces the LanguagE-Action Reward Network~(LEARN) based on natural language command from humans, and LEARN captures the correlation between action and human command. The authors define MDP(+L) as a variant of MDP, denoted as $\langle \mathcal{S},\mathcal{A},P,R,\gamma,l \rangle$, where $l$ represents a human-defined language command describing the desired agent behavior, while the other components remain consistent with the elements in MDP. The original reward function in MDP is labeled as $R_{ext}$, and the dense reward determined by the language command $l$ is denoted as $R_{lan}$. To assess whether the agent is following the language command $l$, LEARN extracts the sequence of past actions $\left( a_1,a_2,...,a_{t-1}\right)$ and transforms it into an action-frequency vector $\textbf{f}$. LEARN takes both $\textbf{f}$ and the natural language command $l$ as inputs and produces a probability distribution indicating the relevance between the action-frequency vector and the natural language command. This distribution measures the correlation between $\textbf{a}$ and $l$, which composes the intrinsic language reward $R_{lan}$. Therefore, the target optimal policy can be generated based on the new reward function $R_{ext}+R_{lan}$.
% During testing, LEARN generates intrinsic reward according to fixed human command and past action sequence, illustrating the task progress aligned with the human command. 
% The intrinsic reward helps with both effective learning and extrinsic explainability of action output towards human command.
The auxiliary reward effectively illustrates the quantitative consideration of accomplishing the task based on human command, presenting extrinsic RL explainability for agent-specific behavior.

\subsection{Learn Mattered Features from Human Interactions}
Learning the significant features directly from human interaction iis an effective way to enhance performance as well as extrinsic explainability.
In Section~\ref{sec::state_explaining}, we discussed the utilization of attention-based techniques to learn important features from input vectors of images or videos. In the context of imitation learning frameworks, there are corresponding approaches to obtain attention. \cite{STATE} categorizes these methods as learning attention from humans, where human trainers provide explicit weight distribution such as gaze information and attention maps. 
This kind of explanation can serve as an additional source of evaluative feedback if the RL agent is able to capture it.
\cite{zhang2020atari} first generate and open the human interaction data with gaze information in Atari games.
\cite{guan2021widening} enhance human attention data by perturbing irrelevant regions. The saliency map serves as human explanation to guide agent effective learning. 
\cite{GAZE2} perceive gaze as probabilistic variables that can be predicted using stochastic units embedded in DNNs. Guided by this idea, they develop a gaze framework that selects important features and estimates the uncertainty of human gaze supervisory signals.
As for enhancing explainability, \cite{GAZE1} employ a visual attention model to train a mapping from images to vehicle control signals, which synchronously generates extrinsic explanations on current state components features towards the actions of agent. 
% The training data of human interaction trajectories are provided by humans, while the attention alignment technique is leveraged to establish the connection between the controller and explanations. 
Meanwhile, from the temporal aspect, many methods find the important states during training to extrinsically explain the task-solving process instead of the internal state features based on human expert demonstrations.
\cite{luss2023local} introduce the concept of meta-state, which encapsulates significant states for task completion based on expert transitions. And the meta-state is obtained from spectral clustering. Concurrently, \cite{liu2023curricular} detect states in expert trajectories as task-specific subgoals by considering the uncertainty of the agent, which is quantified through the variance of the critic value on expert transitions.  
Although the agent itself is not explainable, the subgoals effectively represent the task process and provide extrinsic RL explainability for agent behaviors.

\subsection{Subtask Scheduling with Human Annotation}
In terms of task explanation, leveraging human annotation on scheduling subtasks can be utilized to guide hierarchical agent training and enhance intrinsic explainability within the hierarchical agent. In order to expedite the task decomposition process, \cite{TASK} incorporate human annotation and demonstration to train a high-level language generator to schedule the low-level policies. The generator is trained using imitation learning and consists of LSTM networks, which take the encoded state~(containing explicit goal) as input and produce natural language instructions as output. Instead of output language instruction, \cite{garg2022lisa} generate a discrete latent representation of primitive skills in long-term task with the clustering method to guide low-level agents with human-annotated trajectories. Meanwhile, \cite{gao2023towards} further improves agent explainability by collaborating with humans in MOBA games, in which the high-level agent learn to generate explainable meta commands from human. These frameworks utilizing natural language instructions facilitate successful task decomposition and exhibit high generalizability to new tasks. 
Furthermore, \cite{xu2022perceiving} propose an approach to decompose tasks by having humans answer yes-or-no questions regarding the task content. 
% These mentioned method provides a comprehensive intrinsic explainability to the tasks. 
These methods leverage various human annotations to guide high-level agent in producing scheduling signals on specific state inputs. These scheduling signals are inherently explainable to humans and are subsequently captured by low-level agents to generate actual behavior.

\section{Challenges and future directions for XRL}
\label{sec::Challenges and future directions for XRL}
Given the early stage of XRL research, there remain uncertainties regarding aspects such as architecture and evaluation metrics. Drawing upon the reviewed literature on XRL, we present several promising directions for future research.

\subsection{Human Knowledge in XRL}
Human knowledge-intergrated XRL, which incorporates human knowledge as raw explanations or resources, further enhances XRL explainability and efficiency, as highlighted in Section~\ref{sec::Human_knowledge_for_RL_paradigm}.
However, obtaining certain types of human prior knowledge can be challenging. For example, annotating massive amounts of data manually can be time-consuming~\cite{garg2022lisa} and gathering expert trajectory data for dangerous tasks can be difficult~\cite{hitomi2020development}. In such cases, only a limited amount of data containing suboptimal human knowledge with varying quality may be available, posing a challenge for XRL agents to acquire high-quality policies and explanations.
To address this challenge, several methods have been proposed to effectively utilize available data. \cite{ghai2021explainable} introduce an explainable active learning approach, which efficiently learns a teacher model from limited human feedback by providing both predictions and explanations to humans. Preference-based RL~(PbRL)~\cite{wirth2017survey} is another approach that uses human preferences to train the agent and has shown success in tasks with limited annotation~\cite{christiano2017deep}. \cite{zhang2023learning} enhance the explainability of PbRL by simultaneously learning the reward function and state importance. They leverage a perturbation analysis method to quantify the learned state importance, enabling high explainability with minimal human annotation.
Future research on XRL should fully utilize human-annotated data to extract human knowledge for high explainability and performance.

\subsection{Evaluation Methods}
Despite discussing the current evaluation methods for XRL in Section~\ref{sec::Explainable_RL_definitions_and_measurement}, there is still a lack of a widely accepted approach within the DRL community. This can be attributed to the fact that XRL approaches are highly task-specific, making it challenging to establish a universal measurement method due to the diverse forms of explanations. Furthermore, the notion of explainability is often treated subjectively in many papers, with claims of explainability lacking mathematical formulas or rigorous analysis to support their assertions. The establishment of evaluation methods would enable the comparison of different approaches and identification of the state-of-the-art techniques. For instance, \cite{shen2021autopreview} propose a software platform for self-driving that facilitates the comparison of various XRL agents in the same driving scenario and evaluates the precision of explanations provided by the XRL agent. However, in addition to XRL performance and explanation precision, legal and ethical aspects must also be considered when devising the evaluation method to ensure real-world applicability.

\subsection{Multi-part Explainability}
The aforementioned XRL approaches, including our categorization work, primarily focus on making only a single component of the RL framework explainable, resulting in partial explainability and improvements in specific areas. However, a crucial challenge is that the remaining parts of the RL framework continue to lack transparency for experts. Tasks of high complexity, such as self-driving, demand comprehensive explainability for enhanced safety. Consequently, reliance on a single explainable component is insufficient and fails to provide convincing explanations.
To address this issue, incorporating multi-part explainability into the MDP process can offer a potential solution for RL agents. One approach involves constructing an integrated method that combines various part-explaining techniques. For instance, \cite{huber2021local} merge global explanations based on strategy summaries with local explanations derived from saliency maps, which respectively correspond to agent model-explaining and state-explaining. However, the diverse structures and limited applicability of different part-explaining methods make the combination process challenging. A possible approach could involve abstracting them at a higher level and subsequently integrating them.

\subsection{Balance of High Explainability and Effective Training}

It is feasible to achieve both effective training and high explainability in RL agents. Explainability is regarded as an additional attribute to agent performance, which typically results in XRL being perceived as requiring more computational resources compared to the conventional RL approach~\cite{ribeiro2016model}. However, contrasting the unexplainable DNN-based agents that exhibit high performance, several researchers have discovered that employing simpler and more explainable agent models can also achieve excellent performance while maintaining high explainability, such as linear models~\cite{simple_RL_1,simple_RL_2} and decision tree~\cite{VIPER, topin2021iterative}. These findings indicate that the trade-off between explainability and performance is not as rigid as initially perceived. Moreover, it is possible to strike a balance between these two factors by exploring alternative techniques~\cite{Survey_2}. For example, the incorporation of sparsity or explainability constraints into the agent policy~\cite{rudin2022interpretable}, can enhance explainability without compromising performance. Therefore, further research is warranted to determine the optimal balance between explainability and well-training.

\section{Conclusion}
\label{sec::Conclusion_and_future_work}
Explainability has attracted increasing attention in the RL community due to practical, safe, and trustworthy concerns. It endows the RL agent with the ability to exhibit a well-grounded behavior and further convince the human participants.  In this comprehensive survey, we introduce unified concept definitions and taxonomies to summarize and correlate a wide variety of recent advanced XRL approaches. The survey first gives an in-depth introduction to the explainability definition and evaluation metric of XRL. Then we further categorize the related XRL approaches into four branches:~(a)~Agent model-explaining methods that directly build the agent model as an explainable box. (b)~Reward-explaining methods that regularize the reward function to be understandable. (c)~State-explaining methods that provide the attention-based explanation of observations. (d)~Task-explaining methods that decompose the task to get multi-stage explainability.

Moreover, it is notable that several XRL methods conversely leverage human knowledge to promote the optimization process of learning agents. We additionally discuss and organize these works into our taxonomy structure, while the other XRL surveys pay little attention to it. 

We hope that this survey can help newcomers and researchers to understand and exploit the existing methods in the growing XRL field, as well as highlight opportunities and challenges for future research.

% \backmatter

% \section*{Acknowledgments}
% This should be a simple paragraph before the References to thank those individuals and institutions who have supported your work on this article.

\bibliography{simp_survey}

\begin{thebibliography}{100}
\providecommand{\url}[1]{#1}
\csname url@samestyle\endcsname
\providecommand{\newblock}{\relax}
\providecommand{\bibinfo}[2]{#2}
\providecommand{\BIBentrySTDinterwordspacing}{\spaceskip=0pt\relax}
\providecommand{\BIBentryALTinterwordstretchfactor}{4}
\providecommand{\BIBentryALTinterwordspacing}{\spaceskip=\fontdimen2\font plus
\BIBentryALTinterwordstretchfactor\fontdimen3\font minus \fontdimen4\font\relax}
\providecommand{\BIBforeignlanguage}[2]{{%
\expandafter\ifx\csname l@#1\endcsname\relax
\typeout{** WARNING: IEEEtran.bst: No hyphenation pattern has been}%
\typeout{** loaded for the language `#1'. Using the pattern for}%
\typeout{** the default language instead.}%
\else
\language=\csname l@#1\endcsname
\fi
#2}}
\providecommand{\BIBdecl}{\relax}
\BIBdecl

\bibitem{RL}
R.~S. Sutton \emph{et~al.}, \emph{Reinforcement learning: An introduction}.\hskip 1em plus 0.5em minus 0.4em\relax MIT press, 2018.

\bibitem{kjellstrom2010tracking}
H.~Kjellstr{\"o}m \emph{et~al.}, ``Tracking people interacting with objects,'' in \emph{CVPR}, 2010.

\bibitem{yampolskiy2012artificial}
R.~V. Yampolskiy \emph{et~al.}, ``Artificial general intelligence and the human mental model,'' in \emph{Singularity Hypotheses}, 2012.

\bibitem{williamson1999mental}
D.~M. Williamson \emph{et~al.}, ``‘mental model’comparison of automated and human scoring,'' \emph{Journal of Educational Measurement}, 1999.

\bibitem{powers2006advisor}
A.~Powers \emph{et~al.}, ``The advisor robot: tracing people's mental model from a robot's physical attributes,'' in \emph{HRI}, 2006.

\bibitem{stauffer2016components}
W.~R. Stauffer \emph{et~al.}, ``Components and characteristics of the dopamine reward utility signal,'' \emph{Journal of Comparative Neurology}, 2016.

\bibitem{bengio2017deep}
Y.~Bengio \emph{et~al.}, \emph{Deep Learning}, 2016.

\bibitem{sze2017efficient}
V.~Sze \emph{et~al.}, ``Efficient processing of deep neural networks: A tutorial and survey,'' \emph{Proc. IEEE‌‌}, 2017.

\bibitem{DQN}
V.~Mnih \emph{et~al.}, ``Playing atari with deep reinforcement learning,'' \emph{arXiv preprint arXiv:1312.5602}, 2013.

\bibitem{PPO}
J.~Schulman and Wothers, ``Proximal policy optimization algorithms,'' \emph{arXiv preprint arXiv:1707.06347}, 2017.

\bibitem{A3C}
V.~Mnih \emph{et~al.}, ``Asynchronous methods for deep reinforcement learning,'' in \emph{ICML}, 2016.

\bibitem{SAC}
T.~Haarnoja \emph{et~al.}, ``Soft actor-critic: Off-policy maximum entropy deep reinforcement learning with a stochastic actor,'' in \emph{ICML}, 2018.

\bibitem{TD3}
S.~Fujimoto \emph{et~al.}, ``Addressing function approximation error in actor-critic methods,'' in \emph{ICML}, 2018.

\bibitem{ansuini2019intrinsic}
A.~Ansuini \emph{et~al.}, ``Intrinsic dimension of data representations in deep neural networks,'' in \emph{NeurIPS}, 2019.

\bibitem{yosinski2014transferable}
J.~Yosinski \emph{et~al.}, ``How transferable are features in deep neural networks?'' in \emph{NeurIPS}, 2014.

\bibitem{goldfeld2018estimating}
Z.~Goldfeld and Bothers, ``Estimating information flow in deep neural networks,'' \emph{arXiv preprint arXiv:1810.05728}, 2018.

\bibitem{AlphaGo}
D.~Silver \emph{et~al.}, ``Mastering the game of go without human knowledge,'' \emph{Nature}, 2017.

\bibitem{Dota}
C.~Berner \emph{et~al.}, ``Dota 2 with large scale deep reinforcement learning,'' \emph{arXiv preprint arXiv:1912.06680}, 2019.

\bibitem{chen2019attention}
Y.~Chen \emph{et~al.}, ``Attention-based hierarchical deep reinforcement learning for lane change behaviors in autonomous driving,'' in \emph{CVPR}, 2019.

\bibitem{liu2023CIA}
S.~Liu \emph{et~al.}, ``Contrastive identity-aware learning for multi-agent value decomposition,'' in \emph{AAAI}, 2023.

\bibitem{qinga2po}
Y.~Qing, S.~Liu, J.~Cong, K.~Chen, Y.~Zhou, and M.~Song, ``A2po: Towards effective offline reinforcement learning from an advantage-aware perspective,'' \emph{Advances in Neural Information Processing Systems}, vol.~37, pp. 29\,064--29\,090, 2024.

\bibitem{qing2025bitrajdiff}
Y.~Qing, S.~Chen, Y.~Chi, S.~Liu, S.~Lin, and C.~Zou, ``Bitrajdiff: Bidirectional trajectory generation with diffusion models for offline reinforcement learning,'' \emph{arXiv preprint arXiv:2506.05762}, 2025.

\bibitem{zhou2023centralized}
Y.~Zhou \emph{et~al.}, ``Is centralized training with decentralized execution framework centralized enough for marl?'' \emph{arXiv preprint arXiv:2305.17352}, 2023.

\bibitem{fayjie2018driverless}
A.~R. Fayjie \emph{et~al.}, ``Driverless car: Autonomous driving using deep reinforcement learning in urban environment,'' in \emph{UR}, 2018.

\bibitem{wang2018deep}
S.~Wang \emph{et~al.}, ``Deep reinforcement learning for autonomous driving,'' \emph{arXiv preprint arXiv:1811.11329}, 2018.

\bibitem{chen2019model}
J.~Chen \emph{et~al.}, ``Model-free deep reinforcement learning for urban autonomous driving,'' in \emph{TITS}, 2019.

\bibitem{hoel2019combining}
C.-J. Hoel \emph{et~al.}, ``Combining planning and deep reinforcement learning in tactical decision making for autonomous driving,'' \emph{TIV}, 2019.

\bibitem{wang2017formulation}
P.~Wang \emph{et~al.}, ``Formulation of deep reinforcement learning architecture toward autonomous driving for on-ramp merge,'' in \emph{ITSC}, 2017.

\bibitem{chen2024powerformer}
K.~Chen \emph{et~al.}, ``Powerformer: A section-adaptive transformer for power flow adjustment,'' \emph{arXiv preprint arXiv:2401.02771}, 2024.

\bibitem{xu2024temporal}
F.~Xu \emph{et~al.}, ``Temporal prototype-aware learning for active voltage control on power distribution networks,'' in \emph{KDD}, 2024.

\bibitem{lin2020deep}
L.~Lin \emph{et~al.}, ``Deep reinforcement learning for economic dispatch of virtual power plant in internet of energy,'' \emph{IEEE Internet of Things Journal}, 2020.

\bibitem{yang2021dynamic}
T.~Yang \emph{et~al.}, ``Dynamic energy dispatch strategy for integrated energy system based on improved deep reinforcement learning,'' \emph{Energy}, 2021.

\bibitem{liu2023MAM}
S.~Liu \emph{et~al.}, ``Transmission interface power flow adjustment: A deep reinforcement learning approach based on multi-task attribution map,'' \emph{IEEE TPS}, 2023.

\bibitem{liu2023PAC}
------, ``Progressive decision-making framework for power system topology control,'' \emph{ESWA}, 2024.

\bibitem{he2016deep}
K.~He \emph{et~al.}, ``Deep residual learning for image recognition,'' in \emph{CVPR}, 2016.

\bibitem{zahavy2016graying}
T.~Zahavy \emph{et~al.}, ``Graying the black box: Understanding dqns,'' in \emph{ICML}, 2016.

\bibitem{jaunet2020drlviz}
T.~Jaunet \emph{et~al.}, ``Drlviz: Understanding decisions and memory in deep reinforcement learning,'' in \emph{CGF}, 2020.

\bibitem{ivchyk2024overcoming}
V.~Ivchyk, ``Overcoming barriers to artificial intelligence adoption,'' \emph{Three Seas Economic Journal}, 2024.

\bibitem{prasetya2020navigation}
I.~Prasetya \emph{et~al.}, ``Navigation and exploration in 3d-game automated play testing,'' in \emph{International Workshop on Automating TEST Case Design, Selection, and Evaluation}, 2020.

\bibitem{han2020improving}
X.~Han \emph{et~al.}, ``Improving multi-agent reinforcement learning with imperfect human knowledge,'' in \emph{ICANN}, 2020.

\bibitem{rosenfeld2018leveraging}
A.~Rosenfeld \emph{et~al.}, ``Leveraging human knowledge in tabular reinforcement learning: A study of human subjects,'' \emph{KER}, 2018.

\bibitem{zhang2019leveraging}
R.~Zhang \emph{et~al.}, ``Leveraging human guidance for deep reinforcement learning tasks,'' \emph{arXiv preprint arXiv:1909.09906}, 2019.

\bibitem{zhang2019faster}
H.~Zhang \emph{et~al.}, ``Faster and safer training by embedding high-level knowledge into deep reinforcement learning,'' \emph{arXiv preprint arXiv:1910.09986}, 2019.

\bibitem{guan2020explanation}
L.~Guan \emph{et~al.}, ``Explanation augmented feedback in human-in-the-loop reinforcement learning,'' \emph{arXiv preprint arXiv:2006.14804}, 2020.

\bibitem{guan2021widening}
------, ``Widening the pipeline in human-guided reinforcement learning with explanation and context-aware data augmentation,'' in \emph{NeurIPS}, 2021.

\bibitem{silva2021encoding}
A.~Silva \emph{et~al.}, ``Encoding human domain knowledge to warm start reinforcement learning,'' in \emph{AAAI}, 2021.

\bibitem{object-saliency-map}
R.~Iyer \emph{et~al.}, ``Transparency and explanation in deep reinforcement learning neural networks,'' in \emph{AIES}, 2018.

\bibitem{perturbation-based-saliency}
S.~Greydanus \emph{et~al.}, ``Visualizing and understanding atari agents,'' in \emph{ICML}, 2018.

\bibitem{STATE}
R.~Zhang \emph{et~al.}, ``Leveraging human guidance for deep reinforcement learning tasks,'' \emph{arXiv preprint arXiv:1909.09906}, 2019.

\bibitem{Ration-learning}
U.~Ehsan \emph{et~al.}, ``Automated rationale generation: a technique for explainable ai and its effects on human perceptions,'' in \emph{IUI}, 2019.

\bibitem{Query-based}
B.~Hayes \emph{et~al.}, ``Improving robot controller transparency through autonomous policy explanation,'' in \emph{HRI}, 2017.

\bibitem{x_f_r_1}
B.~RichardWebster \emph{et~al.}, ``Visual psychophysics for making face recognition algorithms more explainable,'' in \emph{ECCV}, 2018.

\bibitem{x_f_r_2}
J.~R. Williford \emph{et~al.}, ``Explainable face recognition,'' in \emph{ECCV}, 2020.

\bibitem{x_f_r_3}
H.~J. andothers, ``Explainable face recognition based on accurate facial compositions,'' in \emph{ICCV}, 2021.

\bibitem{x_f_r_4}
D.~Franco \emph{et~al.}, ``Deep fair models for complex data: Graphs labeling and explainable face recognition,'' \emph{Neurocomputing}, 2022.

\bibitem{x_nlp_1}
H.~Liu \emph{et~al.}, ``Towards explainable nlp: A generative explanation framework for text classification,'' \emph{arXiv preprint arXiv:1811.00196}, 2018.

\bibitem{x_nlp_3}
I.~Arous \emph{et~al.}, ``Marta: Leveraging human rationales for explainable text classification,'' in \emph{AAAI}, 2021.

\bibitem{x_nlp_4}
B.~{\v{S}}krlj \emph{et~al.}, ``autobot: evolving neuro-symbolic representations for explainable low resource text classification,'' \emph{ML}, 2021.

\bibitem{Survey_2}
C.~Glanois \emph{et~al.}, ``A survey on interpretable reinforcement learning,'' \emph{arXiv preprint arXiv:2112.13112}, 2021.

\bibitem{XRL_3}
G.~A. Vouros, ``Explainable deep reinforcement learning: State of the art and challenges,'' \emph{CSUR}, 2022.

\bibitem{heuillet2021explainability}
A.~Heuillet \emph{et~al.}, ``Explainability in deep reinforcement learning,'' \emph{KBS}, 2021.

\bibitem{Survey_4}
L.~Wells \emph{et~al.}, ``Explainable ai and reinforcement learning—a systematic review of current approaches and trends,'' \emph{Frontiers in Artificial Intelligence}, 2021.

\bibitem{broad-xai}
R.~Dazeley \emph{et~al.}, ``Explainable reinforcement learning for broad-xai: A conceptual framework and survey,'' \emph{arXiv preprint arXiv:2108.09003}, 2021.

\bibitem{self-interpretable-2}
E.~Puiutta \emph{et~al.}, ``Explainable reinforcement learning: A survey,'' in \emph{CD-MAKE}, 2020.

\bibitem{milani2023explainable}
S.~Milani \emph{et~al.}, ``Explainable reinforcement learning: A survey and comparative review,'' \emph{CSUR}, 2023.

\bibitem{LMUT}
G.~Liu \emph{et~al.}, ``Toward interpretable deep reinforcement learning with linear model u-trees,'' in \emph{ECML PKDD}, 2018.

\bibitem{AIM}
P.~Madumal \emph{et~al.}, ``Explainable reinforcement learning through a causal lens,'' in \emph{AAAI}, 2020.

\bibitem{verma2019imitation}
A.~Verma \emph{et~al.}, ``Imitation-projected programmatic reinforcement learning,'' in \emph{NeurIPS}, 2019.

\bibitem{garg2022lisa}
D.~Garg \emph{et~al.}, ``Lisa: Learning interpretable skill abstractions from language,'' in \emph{NeurIPS}, 2022.

\bibitem{gao2023towards}
Y.~Gao \emph{et~al.}, ``Towards effective and interpretable human-agent collaboration in moba games: A communication perspective,'' \emph{arXiv preprint arXiv:2304.11632}, 2023.

\bibitem{MDP}
A.~Feinberg, ``Markov decision processes: Discrete stochastic dynamic programming (martin l. puterman),'' \emph{SIAM Review}, 1996.

\bibitem{Double-DQN}
H.~Van~Hasselt \emph{et~al.}, ``Deep reinforcement learning with double q-learning,'' in \emph{AAAI}, 2016.

\bibitem{Dueling-DQN}
Z.~Wang \emph{et~al.}, ``Dueling network architectures for deep reinforcement learning,'' in \emph{ICML}, 2016.

\bibitem{DQN-modify}
V.~Mnih \emph{et~al.}, ``Human-level control through deep reinforcement learning,'' \emph{Nature}, 2015.

\bibitem{DQN-POMDP}
M.~Hausknecht \emph{et~al.}, ``Deep recurrent q-learning for partially observable mdps,'' in \emph{AAAI}, 2015.

\bibitem{PER}
T.~Schaul \emph{et~al.}, ``Prioritized experience replay,'' \emph{arXiv preprint arXiv:1511.05952}, 2015.

\bibitem{Distributional-DQN}
M.~G. Bellemare and Dothers, ``A distributional perspective on reinforcement learning,'' in \emph{ICML}, 2017.

\bibitem{Rainbow-DQN}
M.~Hessel \emph{et~al.}, ``Rainbow: Combining improvements in deep reinforcement learning,'' in \emph{AAAI}, 2018.

\bibitem{ES_DQN}
Q.~Chen \emph{et~al.}, ``Es-dqn: A learning method for vehicle intelligent speed control strategy under uncertain cut-in scenario,'' \emph{TVT}, 2022.

\bibitem{DQN_update}
L.~Meng \emph{et~al.}, ``Improving the diversity of bootstrapped dqn via noisy priors,'' \emph{arXiv preprint arXiv:2203.01004}, 2022.

\bibitem{DQN_update1}
A.~Chraibi \emph{et~al.}, ``Makespan optimisation in cloudlet scheduling with improved {DQN} algorithm in cloud computing,'' \emph{Scientific Programming}, 2021.

\bibitem{DQN_use2}
L.~Chen \emph{et~al.}, ``Conditional dqn-based motion planning with fuzzy logic for autonomous driving,'' \emph{TITS}, 2022.

\bibitem{DQN_use3}
C.~Liu \emph{et~al.}, ``Forecasting the market with machine learning algorithms: An application of {NMC-BERT-LSTM-DQN-X} algorithm in quantitative trading,'' \emph{KDD}, 2022.

\bibitem{DQN_use4}
S.~Park \emph{et~al.}, ``Applying {DQN} solutions in fog-based vehicular networks: Scheduling, caching, and collision control,'' \emph{Vehicular Communications}, 2022.

\bibitem{DQN_use5}
A.~Vashist \emph{et~al.}, ``Dqn based exit selection in multi-exit deep neural networks for applications targeting situation awareness,'' in \emph{ICCE}, 2022.

\bibitem{TRPO}
J.~Schulman \emph{et~al.}, ``Trust region policy optimization,'' in \emph{ICML}, 2015.

\bibitem{DPPO}
N.~Heess \emph{et~al.}, ``Emergence of locomotion behaviours in rich environments,'' \emph{arXiv preprint arXiv:1707.02286}, 2017.

\bibitem{GAE}
J.~Schulman \emph{et~al.}, ``High-dimensional continuous control using generalized advantage estimation,'' \emph{arXiv preprint arXiv:1506.02438}, 2015.

\bibitem{AC_modify1}
B.~Fern{\'{a}}ndez{-}Gauna \emph{et~al.}, ``Actor-critic continuous state reinforcement learning for wind-turbine control robust optimization,'' \emph{Information Sciences}, 2022.

\bibitem{AC_modify2}
X.~Gong \emph{et~al.}, ``Actor-critic with familiarity-based trajectory experience replay,'' \emph{Information Sciences}, vol. 582, pp. 633--647, 2022.

\bibitem{AC_modify3}
X.~Xin \emph{et~al.}, ``Supervised advantage actor-critic for recommender systems,'' in \emph{CIKM}, 2022.

\bibitem{DPG}
D.~Silver \emph{et~al.}, ``Deterministic policy gradient algorithms,'' in \emph{ICML}, 2014.

\bibitem{DDPG}
T.~P. Lillicrap \emph{et~al.}, ``Continuous control with deep reinforcement learning,'' \emph{arXiv preprint arXiv:1509.02971}, 2015.

\bibitem{XRL_1}
T.~Miller, ``Explanation in artificial intelligence: Insights from the social sciences,'' \emph{AI}, 2019.

\bibitem{XRL_2}
B.~Kim \emph{et~al.}, ``Examples are not enough, learn to criticize! criticism for interpretability,'' in \emph{NeurIPS}, 2016.

\bibitem{Decision_tree}
A.~Silva \emph{et~al.}, ``Optimization methods for interpretable differentiable decision trees applied to reinforcement learning,'' in \emph{AISTATS}, 2020.

\bibitem{STG}
T.~Shu \emph{et~al.}, ``Hierarchical and interpretable skill acquisition in multi-task reinforcement learning,'' in \emph{ICLR}, 2018.

\bibitem{NLRL_1}
Z.~Jiang \emph{et~al.}, ``Neural logic reinforcement learning,'' in \emph{ICML}, 2019.

\bibitem{huber2021local}
T.~Huber \emph{et~al.}, ``Local and global explanations of agent behavior: Integrating strategy summaries with saliency maps,'' \emph{AI}, 2021.

\bibitem{leurent2019social}
E.~Leurent \emph{et~al.}, ``Social attention for autonomous decision-making in dense traffic,'' \emph{arXiv preprint arXiv:1911.12250}, 2019.

\bibitem{Interpretability_evaluation}
F.~Doshi-Velez \emph{et~al.}, ``A roadmap for a rigorous science of interpretability,'' \emph{arXiv preprint arXiv:1702.08608}, 2017.

\bibitem{Metrics_for_XAI}
R.~R. Hoffman \emph{et~al.}, ``Metrics for explainable ai: Challenges and prospects,'' \emph{arXiv preprint arXiv:1812.04608}, 2018.

\bibitem{Metrics_for_XAI_plus}
S.~Mohseni \emph{et~al.}, ``A multidisciplinary survey and framework for design and evaluation of explainable ai systems,'' \emph{TiiS‌‌}, 2021.

\bibitem{kay2016ish}
M.~Kay \emph{et~al.}, ``When (ish) is my bus? user-centered visualizations of uncertainty in everyday, mobile predictive systems,'' in \emph{CHI}, 2016.

\bibitem{ribeiro2016should}
M.~T. Ribeiro \emph{et~al.}, ``" why should i trust you?" explaining the predictions of any classifier,'' in \emph{KDD}, 2016.

\bibitem{ribeiro2018anchors}
T.~Ribeiro \emph{et~al.}, ``Anchors: High-precision model-agnostic explanations,'' in \emph{AAAI}, 2018.

\bibitem{nushi2018towards}
B.~Nushi, E.~Kamar, and E.~Horvitz, ``Towards accountable ai: Hybrid human-machine analyses for characterizing system failure,'' in \emph{HCOMP}, 2018.

\bibitem{bansal2019beyond}
G.~Bansal \emph{et~al.}, ``Beyond accuracy: The role of mental models in human-ai team performance,'' in \emph{HCOMP}, 2019.

\bibitem{kim2018interpretability}
B.~Kim \emph{et~al.}, ``Interpretability beyond feature attribution: Quantitative testing with concept activation vectors (tcav),'' in \emph{ICML}, 2018.

\bibitem{kulesza2013too}
T.~Kulesza \emph{et~al.}, ``Too much, too little, or just right? ways explanations impact end users' mental models,'' in \emph{VL/HCC}, 2013.

\bibitem{lakkaraju2016interpretable}
H.~Lakkaraju \emph{et~al.}, ``Interpretable decision sets: A joint framework for description and prediction,'' in \emph{KDD}, 2016.

\bibitem{rader2015understanding}
E.~Rader \emph{et~al.}, ``Understanding user beliefs about algorithmic curation in the facebook news feed,'' in \emph{CHI}, 2015.

\bibitem{bilgic2005explaining}
M.~Bilgic \emph{et~al.}, ``Explaining recommendations: Satisfaction vs. promotion,'' in \emph{IUI Workshop}, 2005.

\bibitem{gedikli2014should}
F.~Gedikli \emph{et~al.}, ``How should i explain? a comparison of different explanation types for recommender systems,'' \emph{IJHCI‌‌}, 2014.

\bibitem{lage2019human}
I.~Lage \emph{et~al.}, ``Human evaluation of models built for interpretability,'' in \emph{HCOMP}, 2019.

\bibitem{lim2009assessing}
B.~Y. Lim \emph{et~al.}, ``Assessing demand for intelligibility in context-aware applications,'' in \emph{ICUC}, 2009.

\bibitem{coppers2018intellingo}
S.~Coppers \emph{et~al.}, ``Intellingo: An intelligible translation environment,'' in \emph{CHI}, 2018.

\bibitem{lim2009and}
B.~Y. Lim \emph{et~al.}, ``Why and why not explanations improve the intelligibility of context-aware intelligent systems,'' in \emph{CHI}, 2009.

\bibitem{berkovsky2017recommend}
S.~Berkovsky \emph{et~al.}, ``How to recommend? user trust factors in movie recommender systems,'' in \emph{IUI}, 2017.

\bibitem{pu2006trust}
P.~Pu and L.~Chen, ``Trust building with explanation interfaces,'' in \emph{IUI}, 2006.

\bibitem{nothdurft2014probabilistic}
F.~Nothdurft \emph{et~al.}, ``Probabilistic human-computer trust handling,'' in \emph{SIGDIAL}, 2014.

\bibitem{RS-rainbow}
Z.~Yang \emph{et~al.}, ``Learn to interpret atari agents,'' \emph{arXiv preprint arXiv:1812.11276}, 2018.

\bibitem{PIRL}
A.~Verma \emph{et~al.}, ``Programmatically interpretable reinforcement learning,'' in \emph{ICML}, 2018.

\bibitem{liu2023curricular}
S.~Liu \emph{et~al.}, ``Curricular subgoals for inverse reinforcement learning,'' \emph{arXiv preprint arXiv:2306.08232}, 2023.

\bibitem{hermans1970questionnaire}
H.~J. Hermans, ``A questionnaire measure of achievement motivation.'' \emph{JAP}, 1970.

\bibitem{lietz2010research}
P.~Lietz, ``Research into questionnaire design: A summary of the literature,'' \emph{IJMR}, 2010.

\bibitem{introspection}
P.~Sequeira \emph{et~al.}, ``Interestingness elements for explainable reinforcement learning: Understanding agents' capabilities and limitations,'' \emph{AI}, 2020.

\bibitem{VIPER}
O.~Bastani \emph{et~al.}, ``Verifiable reinforcement learning via policy extraction,'' in \emph{NeurIPS}, 2018.

\bibitem{tang2021sensory}
Y.~Tang \emph{et~al.}, ``The sensory neuron as a transformer: Permutation-invariant neural networks for reinforcement learning,'' in \emph{NeurIPS}, 2021.

\bibitem{topin2021iterative}
N.~Topin \emph{et~al.}, ``Iterative bounding mdps: Learning interpretable policies via non-interpretable methods,'' in \emph{AAAI}, 2021.

\bibitem{liu2023fidelity}
X.~Liu \emph{et~al.}, ``Fidelity-induced interpretable policy extraction for reinforcement learning,'' \emph{arXiv preprint arXiv:2309.06097}, 2023.

\bibitem{guo2021edge}
W.~Guo \emph{et~al.}, ``Edge: Explaining deep reinforcement learning policies,'' in \emph{NeurIPS}, 2021.

\bibitem{kindermans2019reliability}
P.-J. Kindermans \emph{et~al.}, ``The (un) reliability of saliency methods,'' in \emph{Explainable AI: Interpreting, Explaining and Visualizing Deep Learning}, 2019.

\bibitem{sundararajan2017axiomatic}
M.~Sundararajan \emph{et~al.}, ``Axiomatic attribution for deep networks,'' in \emph{ICML}, 2017.

\bibitem{binder2016analyzing}
A.~Binder \emph{et~al.}, ``Analyzing and validating neural networks predictions,'' in \emph{ICML Workshop}, 2016.

\bibitem{nguyen2020model}
T.~T. Nguyen \emph{et~al.}, ``A model-agnostic approach to quantifying the informativeness of explanation methods for time series classification,'' in \emph{ECML PKDD Workshop}, 2020.

\bibitem{adebayo2018local}
J.~Adebayo \emph{et~al.}, ``Local explanation methods for deep neural networks lack sensitivity to parameter values,'' \emph{arXiv preprint arXiv:1810.03307}, 2018.

\bibitem{adebayo2018sanity}
------, ``Sanity checks for saliency maps,'' in \emph{NeurIPS}, 2018.

\bibitem{gevrey2003review}
M.~Gevrey \emph{et~al.}, ``Review and comparison of methods to study the contribution of variables in artificial neural network models,'' \emph{Ecol. Modell.}, 2003.

\bibitem{Formula_base}
F.~Maes \emph{et~al.}, ``Policy search in a space of simple closed-form formulas: Towards interpretability of reinforcement learning,'' in \emph{DS}, 2012.

\bibitem{inala2020neurosymbolic}
J.~P. Inala \emph{et~al.}, ``Neurosymbolic transformers for multi-agent communication,'' in \emph{NeurIPS}, 2020.

\bibitem{trivedi2021learning}
D.~Trivedi \emph{et~al.}, ``Learning to synthesize programs as interpretable and generalizable policies,'' in \emph{NeurIPS}, 2021.

\bibitem{Formula_2}
D.~Hein \emph{et~al.}, ``Interpretable policies for reinforcement learning by genetic programming,'' \emph{EAAI}, 2018.

\bibitem{Formula_3}
------, ``Generating interpretable reinforcement learning policies using genetic programming,'' in \emph{GECCO}, 2019.

\bibitem{landajuela2021discovering}
M.~Landajuela \emph{et~al.}, ``Discovering symbolic policies with deep reinforcement learning,'' in \emph{ICML}, 2021.

\bibitem{delfosse2024interpretable}
Q.~Delfosse \emph{et~al.}, ``Interpretable and explainable logical policies via neurally guided symbolic abstraction,'' \emph{NeurIPS}, 2024.

\bibitem{Fuzzy_controller_2}
D.~Hein \emph{et~al.}, ``Particle swarm optimization for generating interpretable fuzzy reinforcement learning policies,'' \emph{EAAI}, 2017.

\bibitem{akrour2019towards}
R.~Akrour \emph{et~al.}, ``Towards reinforcement learning of human readable policies,'' in \emph{ECML PKDD workshop}, 2019.

\bibitem{ou2023fuzzy}
L.~Ou \emph{et~al.}, ``Fuzzy centered explainable network for reinforcement learning,'' \emph{TFS}, 2023.

\bibitem{NLRL_2}
A.~Payani \emph{et~al.}, ``Inductive logic programming via differentiable deep neural logic networks,'' \emph{arXiv preprint arXiv:1906.03523}, 2019.

\bibitem{NLRL_3}
------, ``Incorporating relational background knowledge into reinforcement learning via differentiable inductive logic programming,'' \emph{arXiv preprint arXiv:2003.10386}, 2020.

\bibitem{DAGGER}
S.~Ross \emph{et~al.}, ``A reduction of imitation learning and structured prediction to no-regret online learning,'' in \emph{AISTATS}, 2011.

\bibitem{topin2019generation}
N.~Topin \emph{et~al.}, ``Generation of policy-level explanations for reinforcement learning,'' in \emph{AAAI}, 2019.

\bibitem{custode2020evolutionary}
L.~L. Custode \emph{et~al.}, ``Evolutionary learning of interpretable decision trees,'' in \emph{IJCAI}, 2022.

\bibitem{Conservative_Q_Improvement}
A.~M. Roth \emph{et~al.}, ``Conservative q-improvement: Reinforcement learning for an interpretable decision-tree policy,'' \emph{arXiv preprint arXiv:1907.01180}, 2019.

\bibitem{milani2022maviper}
S.~Milani \emph{et~al.}, ``Maviper: Learning decision tree policies for interpretable multi-agent reinforcement learning,'' \emph{arXiv preprint arXiv:2205.12449}, 2022.

\bibitem{self-interpretable-1}
M.~Du \emph{et~al.}, ``Techniques for interpretable machine learning,'' \emph{Communications of the ACM}, 2019.

\bibitem{CUT}
W.~T.~B. Uther \emph{et~al.}, ``Tree based discretization for continuous state space reinforcement learning,'' in \emph{AAAI}, 1998.

\bibitem{Bayesian}
J.~Snoek \emph{et~al.}, ``Practical bayesian optimization of machine learning algorithms,'' in \emph{NeurIPS}, 2012.

\bibitem{lu2016using}
Q.~Lu \emph{et~al.}, ``Using genetic programming with prior formula knowledge to solve symbolic regression problem,'' \emph{Computational Intelligence and Neuroscience}, vol. 2016, 2016.

\bibitem{barwise1977introduction}
J.~Barwise, ``An introduction to first-order logic,'' in \emph{Studies in Logic and the Foundations of Mathematics}, 1977, vol.~90, pp. 5--46.

\bibitem{zimmer2021differentiable}
M.~Zimmer \emph{et~al.}, ``Differentiable logic machines,'' \emph{arXiv preprint arXiv:2102.11529}, 2021.

\bibitem{distillation}
G.~Hinton, O.~Vinyals, J.~Dean \emph{et~al.}, ``Distilling the knowledge in a neural network,'' \emph{arXiv preprint arXiv:1503.02531}, 2015.

\bibitem{olson2021counterfactual}
M.~L. Olson \emph{et~al.}, ``Counterfactual state explanations for reinforcement learning agents via generative deep learning,'' \emph{AI}, 2021.

\bibitem{stein2021generating}
G.~Stein, ``Generating high-quality explanations for navigation in partially-revealed environments,'' in \emph{NeurIPS}, 2021.

\bibitem{amitai2022don}
Y.~Amitai and O.~Amir, ``“i don’t think so”: Summarizing policy disagreements for agent comparison,'' in \emph{AAAI}, 2022.

\bibitem{yu2023explainable}
Z.~Yu \emph{et~al.}, ``Explainable reinforcement learning via a causal world model,'' \emph{arXiv preprint arXiv:2305.02749}, 2023.

\bibitem{meulemans2024would}
A.~Meulemans \emph{et~al.}, ``Would i have gotten that reward? long-term credit assignment by counterfactual contribution analysis,'' \emph{NeurIPS}, 2024.

\bibitem{amitai2024explaining}
Y.~Amitai \emph{et~al.}, ``Explaining reinforcement learning agents through counterfactual action outcomes,'' in \emph{AAAI}, 2024.

\bibitem{boggess2022toward}
K.~Boggess \emph{et~al.}, ``Toward policy explanations for multi-agent reinforcement learning,'' \emph{arXiv preprint arXiv:2204.12568}, 2022.

\bibitem{boggess2023explainable}
------, ``Explainable multi-agent reinforcement learning for temporal queries,'' \emph{arXiv preprint arXiv:2305.10378}, 2023.

\bibitem{verify}
Y.~Kazak \emph{et~al.}, ``Verifying deep-rl-driven systems,'' in \emph{SIGCOMM workshop}, 2019.

\bibitem{zhu2019inductive}
H.~Zhu \emph{et~al.}, ``An inductive synthesis framework for verifiable reinforcement learning,'' in \emph{PLDI}, 2019.

\bibitem{anderson2020neurosymbolic}
G.~Anderson \emph{et~al.}, ``Neurosymbolic reinforcement learning with formally verified exploration,'' in \emph{NeurIPS}, 2020.

\bibitem{jin2022trainify}
P.~Jin \emph{et~al.}, ``Trainify: A cegar-driven training and verification framework for safe deep reinforcement learning,'' in \emph{CAV}, 2022.

\bibitem{IBE_1}
Y.~Fukuchi \emph{et~al.}, ``Autonomous self-explanation of behavior for interactive reinforcement learning agents,'' in \emph{HAI}, 2017.

\bibitem{IBE_2}
------, ``Application of instruction-based behavior explanation to a reinforcement learning agent with changing policy,'' in \emph{NeurIPS}, 2017.

\bibitem{reward-Decomposition}
Z.~Juozapaitis \emph{et~al.}, ``Explainable reinforcement learning via reward decomposition,'' in \emph{IJCAI}, 2019.

\bibitem{liu2023visual}
M.~Liu \emph{et~al.}, ``Visual imitation learning with patch rewards,'' \emph{arXiv preprint arXiv:2302.00965}, 2023.

\bibitem{foerster2018counterfactual}
J.~Foerster \emph{et~al.}, ``Counterfactual multi-agent policy gradients,'' in \emph{AAAI}, 2018.

\bibitem{shapley_Q-value}
J.~Wang \emph{et~al.}, ``Shapley q-value: A local reward approach to solve global reward games,'' in \emph{AAAI}, 2020.

\bibitem{li2021shapley}
J.~Li \emph{et~al.}, ``Shapley counterfactual credits for multi-agent reinforcement learning,'' in \emph{KDD}, 2021.

\bibitem{RARE}
A.~Tabrez \emph{et~al.}, ``Improving human-robot interaction through explainable reinforcement learning,'' in \emph{HRI}, 2019.

\bibitem{reward-level}
D.~Lyu \emph{et~al.}, ``Sdrl: interpretable and data-efficient deep reinforcement learning leveraging symbolic planning,'' in \emph{AAAI}, 2019.

\bibitem{wu2020tree}
J.~Wu \emph{et~al.}, ``Tree-structured policy based progressive reinforcement learning for temporally language grounding in video,'' in \emph{AAAI}, 2020.

\bibitem{wu2021self}
H.~Wu \emph{et~al.}, ``Self-supervised attention-aware reinforcement learning,'' in \emph{AAAI}, 2021.

\bibitem{mirchandani2021ella}
S.~Mirchandani \emph{et~al.}, ``Ella: Exploration through learned language abstraction,'' in \emph{NeurIPS}, 2021.

\bibitem{jin2022creativity}
M.~Jin \emph{et~al.}, ``Creativity of ai: Automatic symbolic option discovery for facilitating deep reinforcement learning,'' in \emph{AAAI}, 2022.

\bibitem{ashwood2022dynamic}
Z.~Ashwood \emph{et~al.}, ``Dynamic inverse reinforcement learning for characterizing animal behavior,'' in \emph{NeurIPS}, 2022.

\bibitem{Shapley_base}
A.~E. Roth, ``Introduction to the shapley value,'' \emph{The Shapley Value}, pp. 1--27, 1988.

\bibitem{SBRL}
J.~Zheng \emph{et~al.}, ``Robust bayesian inverse reinforcement learning with sparse behavior noise,'' in \emph{AAAI}, 2014.

\bibitem{V-SBRL}
I.~Mishra \emph{et~al.}, ``Visual sparse bayesian reinforcement learning: a framework for interpreting what an agent has learned,'' in \emph{SSCI}, 2018.

\bibitem{shapley_his}
K.~Zhang \emph{et~al.}, ``Explainable ai in deep reinforcement learning models for power system emergency control,'' \emph{TCSS}, 2021.

\bibitem{heuillet2022collective}
A.~Heuillet \emph{et~al.}, ``Collective explainable ai: Explaining cooperative strategies and agent contribution in multiagent reinforcement learning with shapley values,'' \emph{IEEE Computational Intelligence Magazine}, 2022.

\bibitem{kenny2022towards}
E.~M. Kenny \emph{et~al.}, ``Towards interpretable deep reinforcement learning with human-friendly prototypes,'' in \emph{ICLR}, 2022.

\bibitem{ragodos2022protox}
R.~Ragodos \emph{et~al.}, ``Protox: Explaining a reinforcement learning agent via prototyping,'' in \emph{NeurIPS}, 2022.

\bibitem{deshmukh2023explaining}
S.~V. Deshmukh \emph{et~al.}, ``Explaining rl decisions with trajectories,'' \emph{arXiv preprint arXiv:2305.04073}, 2023.

\bibitem{sun2024accountability}
H.~Sun \emph{et~al.}, ``Accountability in offline reinforcement learning: Explaining decisions with a corpus of examples,'' \emph{NeurIPS}, 2024.

\bibitem{goel2018unsupervised}
V.~Goel \emph{et~al.}, ``Unsupervised video object segmentation for deep reinforcement learning,'' in \emph{NeurIPS}, 2018.

\bibitem{petsiuk2018rise}
V.~Petsiuk \emph{et~al.}, ``Rise: Randomized input sampling for explanation of black-box models,'' in \emph{BMVC}, 2018.

\bibitem{DQNViz}
J.~Wang \emph{et~al.}, ``Dqnviz: {A} visual analytics approach to understand deep q-networks,'' \emph{TVCG}, 2019.

\bibitem{annasamy2019towards}
R.~M. Annasamy \emph{et~al.}, ``Towards better interpretability in deep q-networks,'' in \emph{AAAI}, 2019.

\bibitem{neuroevolution}
Y.~Tang \emph{et~al.}, ``Neuroevolution of self-interpretable agents,'' in \emph{GECCO}, 2020.

\bibitem{xu2020deep}
Y.~Xu \emph{et~al.}, ``Deep reinforcement learning with stacked hierarchical attention for text-based games,'' in \emph{NeurIPS}, 2020.

\bibitem{pan2020xgail}
M.~Pan \emph{et~al.}, ``xgail: Explainable generative adversarial imitation learning for explainable human decision analysis,'' in \emph{KDD}, 2020.

\bibitem{guo2021machine}
S.~S. Guo \emph{et~al.}, ``Machine versus human attention in deep reinforcement learning tasks,'' in \emph{NeurIPS}, 2021.

\bibitem{waldchen2022training}
S.~W{\"a}ldchen \emph{et~al.}, ``Training characteristic functions with reinforcement learning: Xai-methods play connect four,'' in \emph{ICML}, 2022.

\bibitem{bertoin2022look}
D.~Bertoin \emph{et~al.}, ``Look where you look! saliency-guided q-networks for generalization in visual reinforcement learning,'' in \emph{NeurIPS}, 2022.

\bibitem{peng2022inherently}
X.~Peng \emph{et~al.}, ``Inherently explainable reinforcement learning in natural language,'' in \emph{NeurIPS}, 2022.

\bibitem{beechey2023explaining}
D.~Beechey \emph{et~al.}, ``Explaining reinforcement learning with shapley values,'' in \emph{ICML}, 2023.

\bibitem{wang2024explainable}
C.~Wang \emph{et~al.}, ``Explainable deep adversarial reinforcement learning approach for robust autonomous driving,'' \emph{TIV}, 2024.

\bibitem{van2018contrastive}
J.~van~der Waa \emph{et~al.}, ``Contrastive explanations for reinforcement learning in terms of expected consequences,'' \emph{arXiv preprint arXiv:1807.08706}, 2018.

\bibitem{SPC}
X.~Pan \emph{et~al.}, ``Semantic predictive control for explainable and efficient policy learning,'' in \emph{ICRA}, 2019.

\bibitem{Monte-Carlo-Dropout}
B.~L{\"u}tjens \emph{et~al.}, ``Safe reinforcement learning with model uncertainty estimates,'' in \emph{ICRA}, 2019.

\bibitem{yau2020did}
H.~Yau \emph{et~al.}, ``What did you think would happen? explaining agent behaviour through intended outcomes,'' in \emph{NeurIPS}, 2020.

\bibitem{lee2020weakly}
L.~Lee \emph{et~al.}, ``Weakly-supervised reinforcement learning for controllable behavior,'' in \emph{NeurIPS}, 2020.

\bibitem{hu2023explainable}
B.~Hu \emph{et~al.}, ``An explainable and robust motion planning and control approach for autonomous vehicle on-ramping merging task using deep reinforcement learning,'' \emph{TTE}, 2023.

\bibitem{lee2024refining}
K.~Lee \emph{et~al.}, ``Refining diffusion planner for reliable behavior synthesis by automatic detection of infeasible plans,'' \emph{NeurIPS}, 2024.

\bibitem{wang2019alphastock}
J.~Wang \emph{et~al.}, ``Alphastock: A buying-winners-and-selling-losers investment strategy using interpretable deep reinforcement attention networks,'' in \emph{KDD}, 2019.

\bibitem{hasselt2010double}
H.~Hasselt, ``Double q-learning,'' in \emph{NeurIPS}, 2010.

\bibitem{bool_task}
G.~Nangue~Tasse \emph{et~al.}, ``A boolean task algebra for reinforcement learning,'' in \emph{NeurIPS}, 2020.

\bibitem{jiang2019language}
Y.~Jiang \emph{et~al.}, ``Language as an abstraction for hierarchical deep reinforcement learning,'' in \emph{NeurIPS}, 2019.

\bibitem{D2D}
B.~Beyret \emph{et~al.}, ``Dot-to-dot: Explainable hierarchical reinforcement learning for robotic manipulation,'' in \emph{IROS}, 2019.

\bibitem{model_primitives}
B.~Wu \emph{et~al.}, ``Model primitives for hierarchical lifelong reinforcement learning,'' \emph{AAMAS}, 2020.

\bibitem{sodhani2021multi}
S.~Sodhani \emph{et~al.}, ``Multi-task reinforcement learning with context-based representations,'' in \emph{ICML}, 2021.

\bibitem{zhong2022improving}
V.~Zhong \emph{et~al.}, ``Improving policy learning via language dynamics distillation,'' in \emph{NeurIPS}, 2022.

\bibitem{luss2023local}
R.~Luss \emph{et~al.}, ``Local explanations for reinforcement learning,'' in \emph{IJCAI}, 2023.

\bibitem{KoGuN}
P.~Zhang \emph{et~al.}, ``Kogun: accelerating deep reinforcement learning via integrating human suboptimal knowledge,'' \emph{arXiv preprint arXiv:2002.07418}, 2020.

\bibitem{LEARN}
P.~Goyal \emph{et~al.}, ``Using natural language for reward shaping in reinforcement learning,'' in \emph{IJCAI}, 2019.

\bibitem{GAZE1}
J.~Kim \emph{et~al.}, ``Textual explanations for self-driving vehicles,'' in \emph{ECCV}, 2018.

\bibitem{GAZE2}
Y.~Li \emph{et~al.}, ``In the eye of beholder: Joint learning of gaze and actions in first person video,'' in \emph{ECCV}, 2018.

\bibitem{TASK}
V.~Chen \emph{et~al.}, ``Ask your humans: Using human instructions to improve generalization in reinforcement learning,'' in \emph{ICLR}, 2021.

\bibitem{rudolf2022fuzzy}
T.~Rudolf \emph{et~al.}, ``Fuzzy action-masked reinforcement learning behavior planning for highly automated driving,'' in \emph{CCAR}, 2022.

\bibitem{shi2022efficient}
W.~Shi \emph{et~al.}, ``Efficient hierarchical policy network with fuzzy rules,'' \emph{IJMLC}, 2022.

\bibitem{ng1999policy}
A.~Y. Ng \emph{et~al.}, ``Policy invariance under reward transformations: Theory and application to reward shaping,'' in \emph{ICML}, 1999.

\bibitem{burda2018exploration}
Y.~Burda \emph{et~al.}, ``Exploration by random network distillation,'' \emph{arXiv preprint arXiv:1810.12894}, 2018.

\bibitem{badia2020agent57}
A.~P. Badia \emph{et~al.}, ``Agent57: Outperforming the atari human benchmark,'' in \emph{ICML}, 2020.

\bibitem{harutyunyan2019hindsight}
A.~Harutyunyan \emph{et~al.}, ``Hindsight credit assignment,'' in \emph{NeurIPS}, 2019.

\bibitem{liu2019sequence}
Y.~Liu \emph{et~al.}, ``Sequence modeling of temporal credit assignment for episodic reinforcement learning,'' \emph{arXiv preprint arXiv:1905.13420}, 2019.

\bibitem{zhang2020atari}
R.~Zhang \emph{et~al.}, ``Atari-head: Atari human eye-tracking and demonstration dataset,'' in \emph{AAAI}, 2020.

\bibitem{xu2022perceiving}
Y.~Xu \emph{et~al.}, ``Perceiving the world: Question-guided reinforcement learning for text-based games,'' \emph{arXiv preprint arXiv:2204.09597}, 2022.

\bibitem{hitomi2020development}
K.~Hitomi \emph{et~al.}, ``Development of a dangerous driving suppression system using inverse reinforcement learning and blockchain,'' in \emph{DCAI}, 2020.

\bibitem{ghai2021explainable}
B.~Ghai \emph{et~al.}, ``Explainable active learning (xal) toward ai explanations as interfaces for machine teachers,'' \emph{HCI}, 2021.

\bibitem{wirth2017survey}
C.~Wirth \emph{et~al.}, ``A survey of preference-based reinforcement learning methods,'' \emph{JMLR}, 2017.

\bibitem{christiano2017deep}
P.~F. Christiano \emph{et~al.}, ``Deep reinforcement learning from human preferences,'' in \emph{NeurIPS}, 2017.

\bibitem{zhang2023learning}
G.~Zhang \emph{et~al.}, ``Learning state importance for preference-based reinforcement learning,'' \emph{ML}, 2023.

\bibitem{shen2021autopreview}
Y.~Shen \emph{et~al.}, ``Autopreview: A framework for autopilot behavior understanding,'' in \emph{CHI}, 2021.

\bibitem{ribeiro2016model}
M.~T. Ribeiro \emph{et~al.}, ``Model-agnostic interpretability of machine learning,'' \emph{arXiv preprint arXiv:1606.05386}, 2016.

\bibitem{simple_RL_1}
H.~Mania \emph{et~al.}, ``Simple random search of static linear policies is competitive for reinforcement learning,'' in \emph{NeurIPS}, 2018.

\bibitem{simple_RL_2}
C.~Rudin \emph{et~al.}, ``The secrets of machine learning: ten things you wish you had known earlier to be more effective at data analysis,'' in \emph{INFORMS}, 2019.

\bibitem{rudin2022interpretable}
------, ``Interpretable machine learning: Fundamental principles and 10 grand challenges,'' \emph{Statistic Surveys}, 2022.

\end{thebibliography}
\bibliographystyle{IEEEtran}

\vfill

\end{document}